\documentclass[10pt,twocolumn,letterpaper]{article}

\usepackage{cvpr}              
\usepackage{multirow}
\usepackage[table]{xcolor}
\usepackage{pifont}
\usepackage{algorithm}
\usepackage{algpseudocode}
\usepackage{cuted}
\usepackage{caption}
\usepackage[accsupp]{axessibility} 

\definecolor{cvprblue}{rgb}{0.21,0.49,0.74}
\usepackage[pagebackref,breaklinks,colorlinks,allcolors=cvprblue]{hyperref}

\def\paperID{****} 
\def\confName{CVPR}
\def\confYear{2026}

\title{TWINGS: Thin Plate Splines Warp-aligned Initialization for Sparse-View Gaussian Splatting}

\author{
    Hyeseong Kim$^{1,2}$ \quad Geonhui Son$^{1}$ \quad Deukhee Lee$^{1,2}$\thanks{Corresponding authors.} \quad Dosik Hwang$^{1,2}$\footnotemark[1] \\ [5pt] 
    $^1$ Yonsei University \quad $^2$ Korea Institute of Science and Technology \\
{\tt\small \{hyeseongkim, higun2, dosik.hwang\}@yonsei.ac.kr} \ {\tt\small dkylee@kist.re.kr}}

\begin{document}
\maketitle

\begin{abstract}
Novel view synthesis from sparse-view inputs poses a significant challenge in 3D computer vision, particularly for achieving high-quality scene reconstructions with limited viewpoints. We introduce TWINGS, a framework that enhances 3D Gaussian Splatting (3DGS) by directly addressing point sparsity. We employ Thin Plate Splines (TPS), a smooth non-rigid deformation model that minimizes bending energy to estimate a globally coherent warp from control-point correspondences, to align backprojected points from estimated depth with triangulated 3D control points, yielding calibrated backprojected points. By sampling these calibrated points near the control points, TWINGS provides a fast and geometrically accurate initialization for 3DGS, ultimately improving structural detail preservation and color fidelity in reconstructed scenes. Extensive experiments on DTU, LLFF, and Mip-NeRF360 demonstrate that TWINGS consistently outperforms existing methods, delivering detailed and accurate reconstructions under sparse-view scenarios. Project page: \url{https://sandokim.github.io/twings/}.

\end{abstract}

\section{Introduction}
\label{sec:intro}

\begin{figure}[!htbp]
    \centering
    \includegraphics[width=\columnwidth]{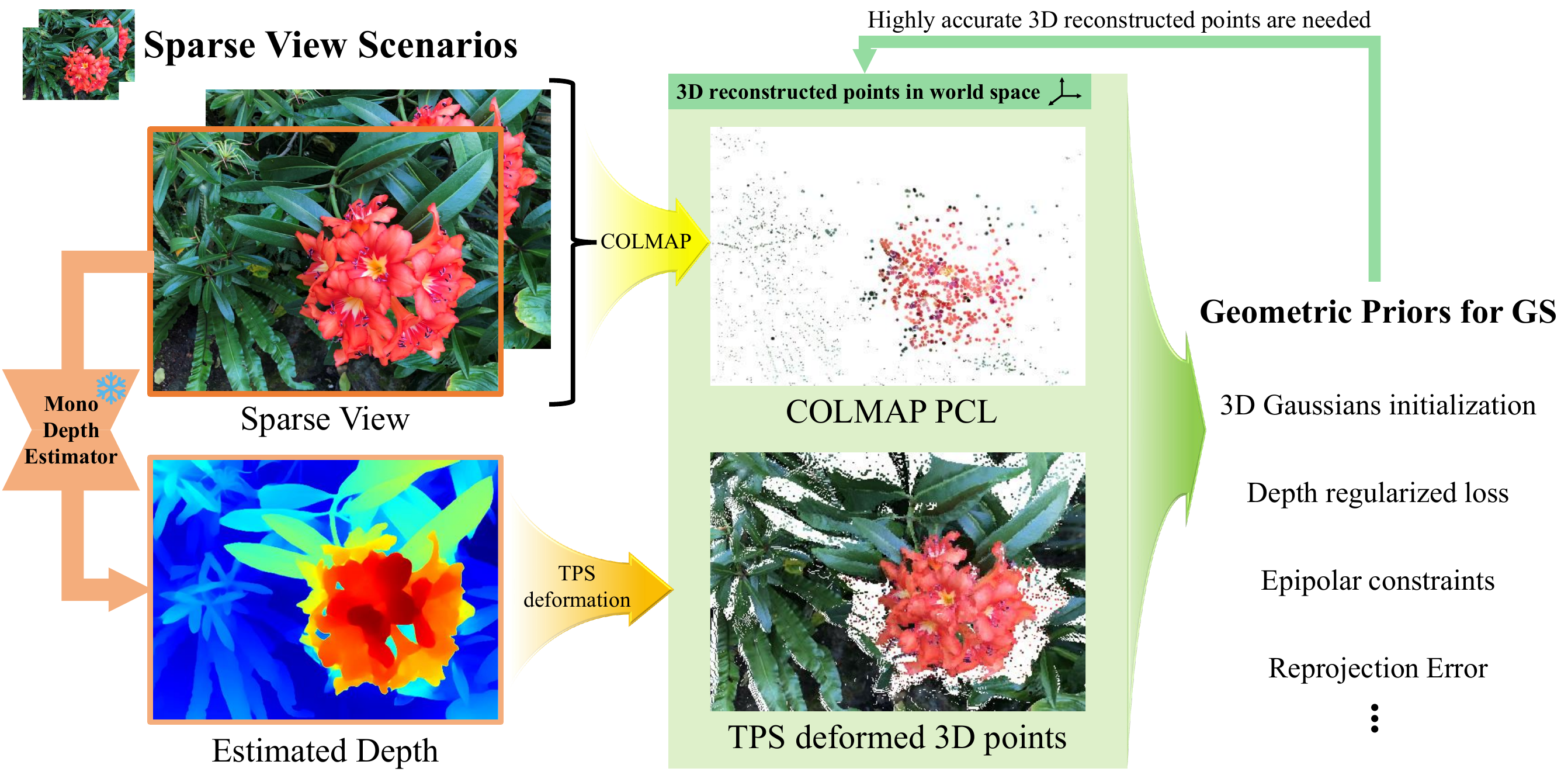}
    \caption{Visualization of 3D reconstructed points. Highly accurate 3D reconstructed points are required to constrain 3DGS.}
    \label{fig:fig1}
\end{figure}

Novel view synthesis from sparse input views has emerged as a critical challenge in 3D computer vision. Recent advances in 3D Gaussian Splatting (3DGS) \cite{kerbl20233d} have demonstrated impressive capabilities for real-time, high-quality rendering. 3DGS represents scenes using 3D Gaussian primitives, which are typically initialized from a point cloud reconstructed via Structure-from-Motion (SfM)~\cite{Schonberger_2016_CVPR}, most notably COLMAP point cloud (PCL). These reconstructed 3D points provide essential geometric constraints that stabilize the optimization of 3DGS, as illustrated in Fig.~\ref{fig:fig1}.

However, under sparse-view scenarios, COLMAP struggles to find sufficient feature correspondences, yielding severely sparse point clouds that lack the geometric cues to constrain the scene, leading 3DGS to degraded reconstruction quality \cite{zhu2023fsgs,zhang2024corgssparseview3dgaussian}. Consequently, 3DGS models trained with such initialization are prone to overfitting the training views, often converging to incorrect local minima or producing distracting floating artifacts \cite{chung2024depth, dropgaussian}. While 3DGS incorporates a densification process~\cite{kerbl20233d}, the extreme sparsity of the initial points hinder its ability to place new Gaussians in geometrically valid locations \cite{zhu2023fsgs}. The result is degraded reconstruction quality, particularly in regions lacking sufficient initial points \cite{zhang2024corgssparseview3dgaussian, zhu2023fsgs, li2024dngaussian, turkulainen2024dn}.

To mitigate this, many recent works have turned to leveraging monocular depth estimates as a geometric prior \cite{li2024dngaussian,turkulainen2024dn,zhu2023fsgs,chung2024depth,comapgs}. Using depth priors, however, introduces the fundamental challenge of inherent scale ambiguity. The problem is particularly acute in sparse-view scenarios, presenting a critical dilemma where the initial point cloud is extremely sparse and unreliable, while the alternative source of geometric information, monocular depth, is fundamentally scale ambiguous, leaving the 3DGS optimization insufficiently constrained.

Previous methods attempt to resolve the ambiguity by aligning monocular depth estimates with the sparse COLMAP PCL, using the result to densify the initial PCL. Typical solutions involve applying a single global scaling factor \cite{chung2024depth} or learning a correlation \cite{comapgs}. Such approaches, being limited to a single rigid transformation, are fundamentally incapable of correcting the complex, non-rigid warping required to truly align the geometry from different views. The central challenge is to warp-align the dense yet geometrically inconsistent pointmap from monocular depth with scene geometry. To address this challenge, we adopt Thin Plate Splines (TPS)~\cite{bookstein1989principal} as the deformation model. Our method applies TPS to backprojected depth to produce a calibrated dense point set that faithfully captures scene structure, and samples near reconstructed geometry to yield a geometrically accurate initialization for 3DGS. This results in balanced foreground and background initialization, which effectively constrains Gaussian optimization in sparse-view scenarios.

Our main contributions are as follows:

\begin{enumerate}
\item \textbf{Fast TPS-based Non-rigid Alignment.} We propose a novel initialization pipeline for 3DGS that leverages Thin Plate Splines (TPS)~\cite{bookstein1989principal} to perform a non-rigid, warp-based alignment of dense, backprojected estimated depth, producing a geometrically accurate dense point clouds. The TPS deformation process is highly efficient and is completed within a few seconds.

\item \textbf{Globally Consistent Multi-view Correspondences.} We construct globally consistent correspondence tracks across all training views. By triangulating the correspondence tracks, our method reconstructs 3D control points that provide strong geometric guidance for the TPS deformation.

\item \textbf{Calibrated Backprojected Points Sampling (CBPS).} Aligned points in the vicinity of reliable control points are selectively sampled to initialize 3DGS.

\item \textbf{Efficient Plug-and-Play Module.} TWINGS-Init is introduced as a plug-and-play initialization module comprising the three components above and integrating seamlessly with existing 3DGS frameworks.
\end{enumerate}

Extensive experiments on the DTU \cite{DTU}, LLFF \cite{llff}, and Mip-NeRF360 \cite{mipnerf360} benchmark datasets demonstrate that TWINGS achieves state-of-the-art performance, outperforming existing methods in extremely sparse-view scenarios with significant improvements in both photometric quality and geometric accuracy.

\section{Related Work}
\label{sec:related}

\subsection{Sparse Novel View Synthesis}
\textbf{NeRF-based Approaches.} The challenge of novel view synthesis (NVS) is significantly amplified under sparse-view conditions. Early efforts to address this primarily focused on Neural Radiance Fields (NeRF) \cite{blender, niemeyer2022regnerf, jain2021putting, chen2021mvsnerf, wang2023sparsenerf, yang2023freenerf}. RegNeRF~\cite{niemeyer2022regnerf} introduces regularization terms to enforce geometric consistency. DietNeRF~\cite{jain2021putting} leverages semantic information. Nevertheless, NeRF-based methods still suffer from expensive per-scene optimization and often yield blurry renderings under sparse-view scenarios.

\textbf{3DGS-based Approaches.} The advent of 3D Gaussian Splatting (3DGS) \cite{kerbl20233d} offered a path to real-time, high-quality rendering, but its reliance on an explicit point-based representation introduced unique challenges for sparse-view NVS. Unlike implicit NeRF models, the quality of a 3DGS reconstruction is directly tied to the quality of initial point cloud from SfM~\cite{Schonberger_2016_CVPR}. With few input views, SfM fails to produce a dense and accurate set of points, causing 3DGS to suffer from severe artifacts, incomplete geometry, and overfitting to training views \cite{zhu2023fsgs,li2024dngaussian}. Consequently, a major focus of recent research has been to develop strategies that make 3DGS robust to this poor geometric initialization. FSGS~\cite{zhu2023fsgs} introduces Proximity-guided Gaussian Unpooling to strategically place new Gaussians between existing ones. CoR-GS~\cite{zhang2024corgssparseview3dgaussian} selectively prunes unstable outliers identified via high point disagreement between two parallel models. DNGaussian~\cite{li2024dngaussian} introduces a depth regularization scheme that leverages monocular depth priors to optimize the positions and opacities of Gaussian primitives during training. DropGaussian~\cite{dropgaussian} employs a random dropout strategy to improve visibility and gradient flow for the remaining Gaussians. CoMapGS~\cite{comapgs} constructs covisibility maps and applies supervision weighted by a proximity classifier. 

In contrast, this paper focuses exclusively on initialization and empirically demonstrates that a well-calibrated initialization alone markedly improves 3DGS performance in sparse-view scenarios.


\subsection{Deformation Fields}
\label{sec:deformation_fields}
A promising strategy for creating a high quality initial point cloud for sparse view 3DGS is to first generate a dense set of points by backprojecting a monocular depth map, and then correct its inherent geometric inconsistencies using a sparse set of triangulated 3D reconstructed points defined in a common world frame. This correction process requires a deformation model capable of warping the entire dense, distorted point cloud to align precisely with these triangulated references.

A straightforward approach is Linear Scaling (LS)~\cite{chung2024depth}, which aligns the monocular depth to the sparse COLMAP depths by fitting a single global scale and offset and then backprojecting the adjusted depth. However, the global scale-offset fit cannot capture spatially varying biases, limiting alignment fidelity. Free-Form Deformation (FFD)~\cite{ffd} uses a control-point lattice around the geometry but does not guarantee that points from the backprojected estimated depth align with triangulated 3D points. NURBS~\cite{nurbs, nurbs_survey} use control points and weights to represent both analytic shapes and complex free-form surfaces, but this flexibility incurs storage and computational overhead and is often inefficient when the goal is to align geometry to a fixed set of triangulated points or correspondences.

In this work, we employ Thin Plate Splines (TPS)~\cite{bookstein1989principal}, which are designed for correspondence-driven alignment. TPS computes a smooth deformation field that exactly interpolates the correspondences between backprojected points and multi-view triangulated 3D points while minimizing bending energy. This property aligns with our goal by enforcing exact agreement at reliable triangulated references and smoothly propagating coherent corrections to neighboring regions, yielding an initial point cloud that is both dense and geometrically consistent. Our method utilizes TPS for its favorable trade-off between performance and computational time, as it provides a geometrically accurate deformation within seconds, making it a practical solution that avoids introducing a significant bottleneck to the 3DGS initialization pipeline.

\begin{figure*}[t]
    \centering
    \includegraphics[width=\textwidth]{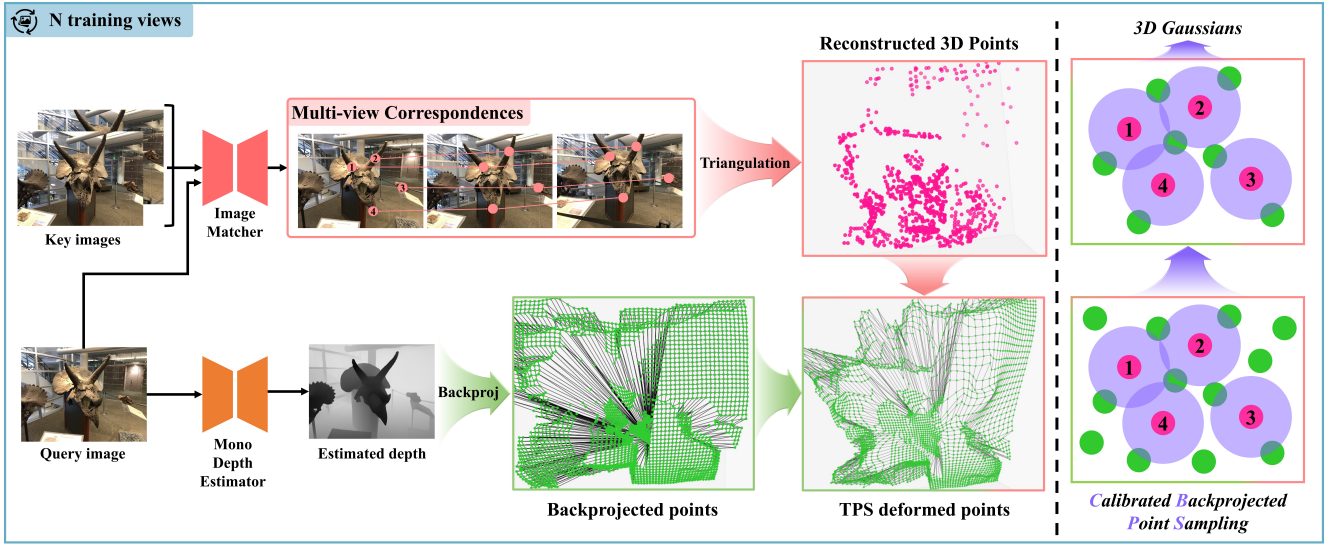}
    \caption{\textbf{TWINGS Pipeline.} Our method consists of three key components: \textbf{Multi-view Correspondences}: We establish multi-view correspondences among query and key images. Using correspondences with known camera parameters, we reconstruct 3D points that correspond to desired control points (pink). By backprojecting the estimated depth, we generate backprojected points (green). \textbf{TPS deformation:} We define a TPS model that deforms the backprojected points to obtain TPS deformed points, which are referred to as CBP. \textbf{CBPS:} The CBP are then strategically sampled near desired control points (pink) to initialize 3DGS training.}
    \label{fig:fig2}
\end{figure*}

\section{Background}
\subsection{Preliminary: 3D Gaussian Splatting}
\label{sec:3dgs}
3D Gaussian Splatting (3DGS) models a 3D scene using a collection \( \mathcal{G} = \{g_i\ | i=1,...,N \} \) of \( N \) differentiable 3D Gaussian primitives. Each Gaussian primitive is parameterized by its mean $\mu \in \mathbb{R}^3$, a covariance matrix $\Sigma \in \mathbb{R}^{3 \times 3}$, which is decomposed into a scaling vector $s \in \mathbb{R}^3$ and a rotation quaternion $q \in \mathbb{R}^4$, an opacity parameter $o \in \mathbb{R}$, and a color $c \in \mathbb{R}^3$ represented using spherical harmonics. The covariance matrix $\Sigma$ of each Gaussian is parameterized by its eigen decomposition:

\begin{equation}
\Sigma = R S S^\top R^\top,
\label{eq:cov}
\end{equation} 

\noindent{where} $S \in \mathbb{R}^3$ is a 3D scale vector containing the square roots of $\Sigma$'s eigenvalues, and $R \in \text{SO}(3)$ is the rotation matrix computed from the quaternion $q$.

In order to optimize the properties of 3D Gaussians to represent the scene, we need to render these Gaussians into images in a differentiable manner. This rendering process involves projecting 3D Gaussians into 2D Gaussians in camera space, sorting them by z-depth, and alpha-compositing them using a discrete volume rendering equation \cite{kerbl20233d}. The resulting pixel color $C$ is given by:
\begin{equation}
C(p) = \sum_{i \in N} c_i \alpha_i T_i, \quad \text{where } T_i = \prod_{j=1}^{i-1} (1 - \alpha_j),
\label{eq:color}
\end{equation}
\noindent{where} $T_i$ represents the accumulated transmittance at pixel $p$, and $\alpha_i$ is the blending coefficient for a Gaussian with center $\mu_i$ in screen space. This differentiable rendering enables optimization of Gaussian properties by minimizing the difference between rendered and ground truth images. 

\subsection{Geometric Priors}
\textbf{Image Matcher.} Reconstructing accurate 3D geometry fundamentally relies on establishing pixel-wise correspondences between images. This field has evolved from classical hand-crafted local features like SIFT \cite{lowe2004distinctive} to learned detectors and descriptors \cite{yi2016lift, detone2018superpoint, brachmann2019neural, sarlin2020superglue}, and more recently to dense matchers \cite{tang2022quadtree, wang2022matchformer, melekhov2019dgc, truong2020glu, edstedt2023dkm, mast3r}. Our method leverages a dense matcher~\cite{mast3r} to find matches.

\section{Method}
We propose TWINGS: Thin Plate Splines Warp-aligned Initialization for Sparse-View Gaussian Splatting. TWINGS reconstructs 3D points through triangulation across multiple views that capture the real scene geometry, which serve as desired control points. For each 2D correspondence, our method retrieves the corresponding estimated depth and backproject it into 3D space to generate initial control points. Our method defines the TPS model that warps the initial control points to align with the desired control points, as illustrated in Fig.~\ref{fig:fig2}. By applying the defined TPS model to the entire set of backprojected points, a dense set of deformed points is obtained that is accurately aligned with the reconstructed scene geometry. These deformed points are referred to as calibrated backprojected points (CBP). Finally, our method samples CBP near the desired control points and combines them with COLMAP PCL to provide a denser and more detailed real scene geometry for 3DGS training.



\subsection{Dense Pointmap Deformation}
\noindent{\textbf{Geometric Alignment.}} As discussed in Sec.~\ref{sec:deformation_fields}, to bridge the gap between reconstructed 3D points and estimated depth by the pretrained monocular depth estimator \cite{DAv2}, we introduce a pointmap and their relationship with multi-view geometry principles \cite{hartley2003multiple}. A pointmap represents a dense 2D field of 3D points, denoted by $X \in \mathbb{R}^{W \times H \times 3}$, where each pixel $(i,j)$ in an RGB image $I_{i,j} \in \mathbb{R}^{W \times H}$ maps to a corresponding 3D point $X_{i,j} \in \mathbb{R}^3$. The mapping from 2D pixel coordinates to 3D points in the camera frame can be computed directly from the depth map $D_{i,j} \in \mathbb{R}^{W \times H}$ and the intrinsic camera matrix $K \in \mathbb{R}^{3 \times 3}$. This is given by:

\begin{equation}
X_{i,j} = (K)^{-1} [i, j, 1]^T D_{i,j},
\label{eq:depth_backproj}
\end{equation}

\noindent{where} $[i, j, 1]^T$ are the homogeneous coordinates of the pixel, and $D_{i,j}$ is the depth at pixel $(i,j)$. This backprojection process defined in Eq.~\ref{eq:depth_backproj} yields the pointmap $X$, providing a geometric representation that maps each image pixel to its corresponding 3D point in the camera coordinate system. Based on this pointmap, a straightforward approach is to initialize 3D Gaussians at each pixel $(i,j)$ by setting their means $\mu \in \mathbb{R}^3$ to $X_{i,j}$ and assigning RGB values $I_{i,j}$ as colors $c \in \mathbb{R}^3$. While the pointmap-based initialization provides a dense set of 3D points, the backprojected points are often mis-scaled with respect to the scene, causing 3DGS training to completely fail. To address this challenge, we propose a deformable registration that aligns backprojected points with reconstructed 3D points derived from triangulation. 

\noindent{\textbf{Multi-view Correspondences.}} Let the training image set be $\mathcal{I}=\{I^{1},\dots,I^{N}\}$,
where $N$ is the number of training views.
For each query image $I^{q}\in\mathcal{I}$,
consider each pixel $p^{q}_i\in I^{q}$ that has matches in other images.
For every such $p^{q}_i$ and every image $I^{j}$ with $j \neq q$,
define the set of matches $C^{j}(p^{q}_i) \subset I^{j}$.
The global set of dense correspondences is constructed as follows:
\begin{equation}
\mathcal{M}
  = \bigcup_{q=1}^{N}
      \bigcup_{p^{q}_i\in I^{q}}
        \Bigl\{\,\bigl(p^{q}_i,\,
            \{\,p^j_{i}\mid j\neq q,\; p^j_{i}\in C^{j}(p^{q}_i) \}\bigr)
        \Bigr\},
\label{eq:matches}
\end{equation}
where each element is a query pixel $p^{q}_i$ paired with the set of pixels
$\{p^{j}_i\}$ from other images that correspond to $p^{q}_i$.

\noindent\textbf{Multi-view Triangulation.}  
For each dense correspondence \(\bigl(p^{q}_i,\{p^{j}_i\}\bigr)\in\mathcal{M}\),  
assume the matched pixels observe the same 3D point \(X_i\).  
Let \(P^{k}\!\in\!\mathbb{R}^{3\times4}\) be the camera projection matrix and \(\tilde p^{k}_{i}\) the homogeneous pixel coordinate in image \(I^{k}_{i}\). The best initial 3D estimate \(\hat{X}_{i}\) is obtained by solving the DLT system of multi-view geometry~\cite{hartley2003multiple}:
\begin{equation}
A X_i = 0,\qquad
A =
\begin{bmatrix}
\tilde p^{q}_i\!\times\! P^{q}\\
\tilde p^{1}_i\!\times\! P^{1}\\
\vdots\\
\tilde p^{K}_i\!\times\! P^{K}
\end{bmatrix},
\label{eq:dlt_system}
\end{equation}
where \(K\) is the number of key images observing the \(i\)-th 3D point. Note that \(K\) may vary across points. $\hat{X_i}$ is then refined by nonlinear optimization to minimize the total reprojection error:
\begin{equation}
X^{*}_i
=\arg\min_{X}
\sum_{k=1}^{K+1}
\bigl\|\pi(P^{k}X)-p^{k}_i\bigr\|_{2}^{2},
\label{eq:reproj_min}
\end{equation}
where \(\pi([x,y,z]^{\top})=[x/z,\; y/z]^{\top}\) denotes the perspective projection. A set of reconstructed 3D points 
$X^{*}=\{X^*_1, \dots,X^{*}_{M}\}$, where $M$ is the total number of correspondences, serves as desired control points for TPS model.

\subsection{Thin Plate Splines for Deformable Registration}
\label{sec:deformable_registration}
\noindent{\textbf{Thin Plate Splines Model Definition.}} Our method utilizes Thin Plate Splines (TPS) \cite{bookstein1989principal} to smoothly deform backprojected points, aligning them with desired control points $X^{*}$. To apply TPS, pairs of initial control points and desired control points are required. Let $D_{est}$ be the estimated depth from query image $I^q$ using the depth estimator \cite{DAv2}. We first obtain $X_{est} \in \mathbb{R}^{W \times H \times 3}$ by backprojecting all pixels of the estimated depth $D_{est}$ using Eq.~\ref{eq:depth_backproj}. For matched pixels $p^q$ identified by the image matcher, we obtain initial control points $X_{est}(p^q) \in \mathbb{R}^{M \times 3}$ by indexing $X_{est}$ at corresponding pixels, while their corresponding desired control points $X^{*}(p^q) \in \mathbb{R}^{M \times 3}$ are obtained by triangulating the dense correspondences in $\mathcal{M}$ associated with $p^q$. For pairs of initial control points and desired control points, the TPS model is defined as follows for each dimension: 

\begin{equation}
\begin{split}
TPS_x = & a_{0x} + a_{xx}x + a_{xy}y + a_{xz}z \\
      & + \sum_{p^q \in M} w_{p^qx} U(\|X_{est}(p^q)-X^{*}(p^q)\|)
\end{split}
\label{eq:TPS_x}
\end{equation}

\begin{equation}
\begin{split}
TPS_y = & a_{0y} + a_{yx}x + a_{yy}y + a_{yz}z \\
      & + \sum_{p^q \in M} w_{p^qy} U(\|X_{est}(p^q)-X^{*}(p^q)\|)
\end{split}
\label{eq:TPS_y}
\end{equation}

\begin{equation}
\begin{split}
TPS_z = & a_{0z} + a_{zx}x + a_{zy}y + a_{zz}z \\
      & + \sum_{p^q \in M} w_{p^qz} U(\|X_{est}(p^q)-X^{*}(p^q)\|)
\end{split}
\label{eq:TPS_z}
\end{equation}

\noindent{The} TPS model consists of a global affine and a local non-affine component. The affine component includes:
\begin{equation}
t = [a_{0x}, a_{0y}, a_{0z}]^T, \quad
A = \begin{bmatrix}
a_{xx} & a_{xy} & a_{xz} \\
a_{yx} & a_{yy} & a_{yz} \\
a_{zx} & a_{zy} & a_{zz},
\end{bmatrix}
\end{equation}

\begin{figure*}[!htbp]
    \centering
    \includegraphics[width=\textwidth]{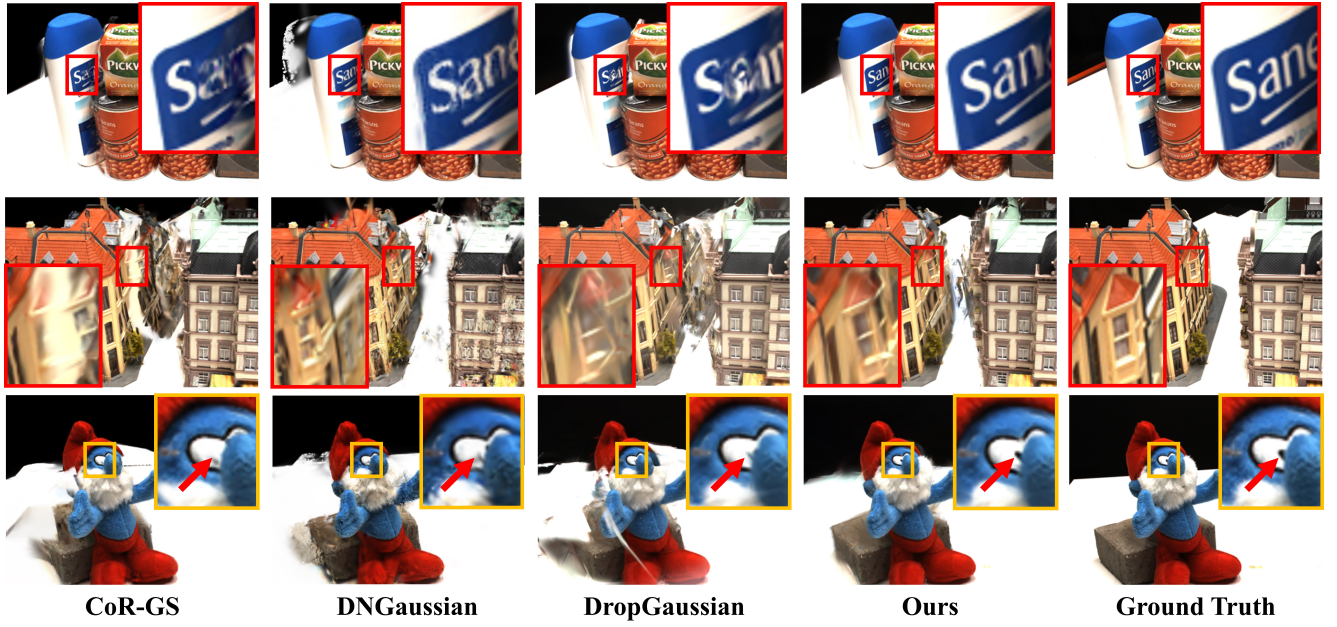}
    \caption{Novel view synthesis results on the DTU dataset~\cite{DTU}. Rendered by CoR-GS~\cite{zhang2024corgssparseview3dgaussian}, DNGaussian~\cite{li2024dngaussian}, DropGaussian~\cite{dropgaussian}, our approach, and ground truth for comparison.}
    \label{fig:dtu_fig}
\end{figure*}

\begin{figure*}[!htbp]
    \centering
    \includegraphics[width=\textwidth]{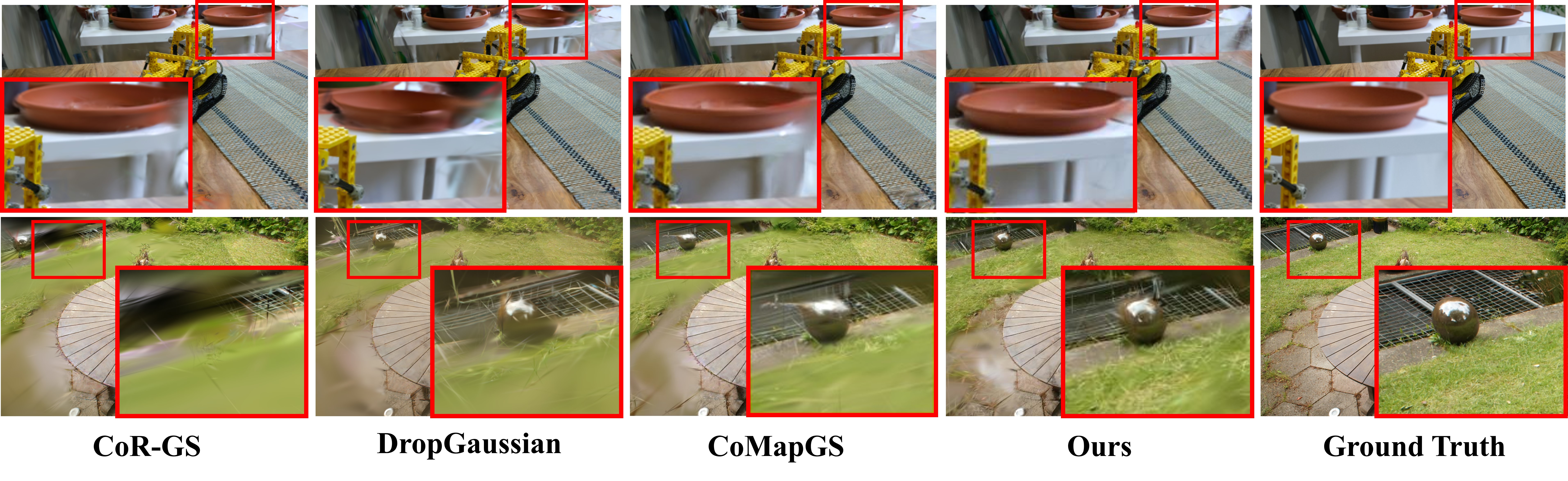}
    \caption{Novel view synthesis results on the Mip-NeRF360 dataset~\cite{mipnerf360}. Rendered by CoR-GS~\cite{zhang2024corgssparseview3dgaussian}, DropGaussian~\cite{dropgaussian}, CoMapGS~\cite{comapgs}, our approach, and ground truth for comparison.}
    \label{fig:mipnerf_fig}
\end{figure*}

\begin{figure*}[!htbp]
    \centering
    \includegraphics[width=\textwidth]{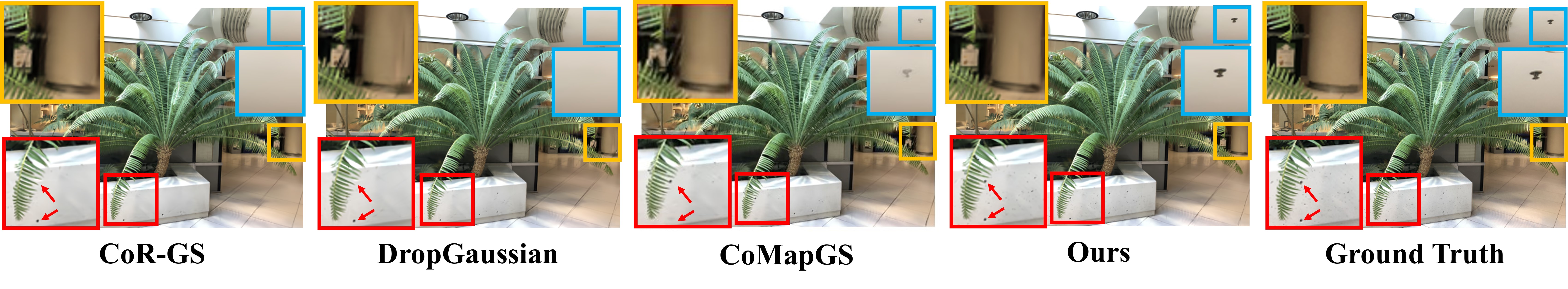}
    \caption{Novel view synthesis results on the LLFF dataset~\cite{llff}. Rendered by CoR-GS~\cite{zhang2024corgssparseview3dgaussian}, DropGaussian~\cite{dropgaussian}, CoMapGS~\cite{comapgs}, our approach, and ground truth for comparison.}
    \label{fig:llff_fig}
\end{figure*}


\noindent{where} $t \in \mathbb{R}^3$ and $A \in \mathbb{R}^{3 \times 3}$ define the global affine component, while the local non-affine component $w_{p^q} \in \mathbb{R}^3$ control the non-linear deformation at each control point. These parameters are determined by solving a linear system derived from Eq. \ref{eq:TPS_x}, Eq. \ref{eq:TPS_y}, and Eq. \ref{eq:TPS_z} using the initial control points $X_{est}(p^q)$ and desired control points $X^{*}(p^q)$. The TPS model can be expressed compactly as:
\begin{equation}
TPS(X) = t + AX + \sum_{p^q \in M} w_{p^q} U(\|X-X^{*}(p^q)\|),
\end{equation}

\noindent{where} $U(r) = r$  a radial basis function that defines how the influence of control points varies with distance, and $w_{p^q}$ is the set of corresponding TPS weights that ensure smooth spatial deformation. TPS is applied to all backprojected points in $X_{est}$, yielding the deformed points $TPS(X_{est})$, which are referred to as calibrated backprojected points (CBP).

\subsection{Calibrated Backprojected Points Sampling}
\label{sec:CBPS}

We aim to initialize 3DGS using 3D points that faithfully represent the real scene geometry.
While the CBP provide accurate initial positions, using all of them does not fully eliminate residual geometric errors.
To obtain a geometry-consistent set, we perform calibrated backprojected points sampling (CBPS)
by selecting CBP near reliable reconstructed 3D points across all training views:
\begin{equation}
\mathcal{S}_{\text{CBPS}}
  = \bigcup_{x \in X^{*}}
      \{\, b \in \mathcal{B} \mid \| b - x \| \le r \,\},
\end{equation}
where $X^{*}$ is the set of reconstructed 3D points, $\mathcal{B}$ is the entire CBP set,
and $r$ is the sampling radius, which is set as a fraction of the scale of the scene $S$, defined as the radius of the bounding sphere enclosing the camera poses~\cite{kerbl20233d, zhang2020nerf++}.
The sampled set $\mathcal{S}_{\text{CBPS}}$ is then used to initialize the 3D Gaussian primitives.

\label{sec:method}

\section{Experiments}

\subsection{Setups}

\begin{table*}[!t]
\centering
\caption{Quantitative results on the DTU dataset with 3, 6, 9 training views. The best, second-best, and third-best results are marked in red, orange, and yellow, respectively. For several scenes in the DTU dataset in the 3-view setting where COLMAP fails, we follow prior work \cite{li2024dngaussian, dropgaussian} and initialize the baselines with random point clouds.}
\label{table:DTU}
\resizebox{\textwidth}{!}{
\begin{tabular}{c|ccc|ccc|ccc|ccc}
\hline
\multirow{2}{*}{\textbf{Method}} & \multicolumn{3}{c|}{\textbf{PSNR$\uparrow$}} & \multicolumn{3}{c|}{\textbf{SSIM$\uparrow$}} & \multicolumn{3}{c|}{\textbf{LPIPS$\downarrow$}} & \multicolumn{3}{c}{\textbf{AVGE$\downarrow$}} \\ \cline{2-13} 
 & \textbf{3-view} & \textbf{6-view} & \textbf{9-view} & \textbf{3-view} & \textbf{6-view} & \textbf{9-view} & \textbf{3-view} & \textbf{6-view} & \textbf{9-view} & \textbf{3-view} & \textbf{6-view} & \textbf{9-view} \\ \hline
DietNeRF \cite{jain2021putting} & 11.85 & 20.63 & 23.83 & 0.633 & 0.778 & 0.823 & 0.314 & 0.201 & 0.173 & 0.243 & 0.101 & 0.068 \\
RegNeRF \cite{niemeyer2022regnerf} & 18.89 & 22.20 & 24.93 & 0.745 & 0.841 & 0.884 & 0.190 & 0.117 & 0.089 & 0.112 & 0.071 & 0.047 \\
Mip-NeRF \cite{barron2021mip} & 9.10 & 16.84 & 23.56 & 0.578 & 0.754 & 0.877 & 0.348 & 0.197 & 0.100 & 0.311 & 0.144 & 0.057 \\
FreeNeRF \cite{yang2023freenerf} & \cellcolor{orange!50}19.92 & 23.25 & 25.38 & 0.787 & 0.844 & 0.888 & 0.182 & 0.137 & 0.096 & 0.098 & 0.068 & 0.046 \\
SparseNeRF \cite{wang2023sparsenerf} & \cellcolor{yellow!50}19.55 & - & - & 0.769 & - & - & 0.201 & - & - & 0.102 & - & - \\
3DGS \cite{kerbl20233d} & 17.65 & 24.00 & 26.85 & 0.816 & 0.907 & 0.942 & 0.146 & 0.076 & 0.049 & 0.108 & 0.050 & 0.032 \\
DNGaussian \cite{li2024dngaussian} & 18.91 & 23.31 & 25.95 & 0.790 & 0.875 & 0.924 & 0.176 & 0.120 & 0.074 & 0.101 & 0.062 & 0.040 \\
FSGS \cite{zhu2023fsgs} & 17.24 & 23.48 & 26.61 & 0.824 & 0.908 & 0.943 & 0.158 & 0.091 & 0.064 & 0.115 & 0.057 & 0.034 \\
CoR-GS \cite{zhang2024corgssparseview3dgaussian} & 19.21 & \cellcolor{orange!50}24.51 & \cellcolor{yellow!50}27.18 & \cellcolor{orange!50}0.853 & \cellcolor{yellow!50}0.917 & \cellcolor{yellow!50}0.947 & \cellcolor{orange!50}0.119 & \cellcolor{orange!50}0.068 & \cellcolor{yellow!50}0.045 & \cellcolor{orange!50}0.087 & \cellcolor{orange!50}0.046 & \cellcolor{yellow!50}0.029 \\
DropGaussian \cite{dropgaussian} & 18.41 & \cellcolor{yellow!50}24.46 & \cellcolor{orange!50}27.75 & \cellcolor{yellow!50}0.849 & \cellcolor{orange!50}0.922 & \cellcolor{orange!50}0.954 & \cellcolor{yellow!50}0.136 & \cellcolor{yellow!50}0.070 & \cellcolor{orange!50}0.042 & \cellcolor{yellow!50}0.099 & \cellcolor{yellow!50}0.048 & \cellcolor{orange!50}0.026 \\
\textbf{TWINGS (Ours)} & \cellcolor{red!50}21.52 & \cellcolor{red!50}25.81 & \cellcolor{red!50}28.22 & \cellcolor{red!50}0.880 & \cellcolor{red!50}0.930 & \cellcolor{red!50}0.958 & \cellcolor{red!50}0.107 & \cellcolor{red!50}0.063 & \cellcolor{red!50}0.040 & \cellcolor{red!50}0.069 & \cellcolor{red!50}0.039 & \cellcolor{red!50}0.024 \\ \hline
\end{tabular}
}
\end{table*}

{
\setlength{\floatsep}{4pt}        
\setlength{\textfloatsep}{6pt}    
\setlength{\intextsep}{6pt}       

\begin{table}[!t]
\caption{Quantitative results on the Mip-NeRF360 dataset with 12, 24 training views. The best, second-best, and third-best results are marked in red, orange, and yellow, respectively.}
\label{table:mipnerf360}
\resizebox{\columnwidth}{!}{
\begin{tabular}{c|cccccc}
\hline
\multirow{2}{*}{\textbf{Method}} & \multicolumn{3}{c}{\textbf{12-view}} & \multicolumn{3}{c}{\textbf{24-view}} \\ \cline{2-7} 
 & PSNR↑ & SSIM↑ & LPIPS↓ & PSNR↑ & SSIM↑ & LPIPS↓ \\ \hline
3DGS \cite{kerbl20233d} & 18.52 & 0.523 & 0.415 & 22.80 & 0.708 & 0.276 \\
FSGS \cite{zhu2023fsgs} & 18.80 & 0.531 & 0.418 & 23.28 & 0.715 & 0.274 \\
CoR-GS \cite{zhang2024corgssparseview3dgaussian} & 19.52 & 0.558 & 0.418 & 23.39 & 0.727 & 0.271 \\
DropGaussian \cite{dropgaussian} & \cellcolor{orange!50}19.74 & \cellcolor{yellow!50}0.577 & \cellcolor{red!50}0.364 & \cellcolor{orange!50}24.13 & \cellcolor{red!50}0.762 & \cellcolor{red!50}0.225 \\
CoMapGS \cite{comapgs} & \cellcolor{yellow!50}19.68 & \cellcolor{orange!50}0.591 & \cellcolor{yellow!50}0.394 & \cellcolor{yellow!50}23.46 & \cellcolor{yellow!50}0.734 & \cellcolor{yellow!50}0.264 \\
\textbf{TWINGS (Ours)} & \cellcolor{red!50}20.35 & \cellcolor{red!50}0.618 & \cellcolor{orange!50}0.368 & \cellcolor{red!50}24.17 & \cellcolor{red!50}0.762 & \cellcolor{orange!50}0.252 \\ \hline
\end{tabular}
}
\end{table}

\begin{table}[!t]
\centering
\caption{Quantitative results on the LLFF dataset with 3, 6, 9 training views. The best, second-best, and third-best results are marked in red, orange, and yellow, respectively.}
\label{table:llff}
\renewcommand{\arraystretch}{0.95} 
\resizebox{\columnwidth}{!}{
\begin{tabular}{c|ccccccccc}
\hline
\multirow{2}{*}{\textbf{Method}} & \multicolumn{3}{c}{\textbf{3-view}} & \multicolumn{3}{c}{\textbf{6-view}} & \multicolumn{3}{c}{\textbf{9-view}} \\ \cline{2-10} 
 & PSNR↑ & SSIM↑ & LPIPS↓ & PSNR↑ & SSIM↑ & LPIPS↓ & PSNR↑ & SSIM↑ & LPIPS↓ \\ \hline
Mip-NeRF \cite{barron2021mip} & 16.10 & 0.401 & 0.460 & 22.91 & 0.756 & 0.213 & 24.88 & 0.826 & 0.160 \\
DietNeRF \cite{jain2021putting} & 14.94 & 0.370 & 0.496 & 21.75 & 0.717 & 0.248 & 24.28 & 0.801 & 0.183 \\
RegNeRF \cite{niemeyer2022regnerf} & 19.08 & 0.587 & 0.336 & 23.10 & 0.760 & 0.206 & 24.86 & 0.820 & 0.161 \\
FreeNeRF \cite{yang2023freenerf} & 19.63 & 0.612 & 0.308 & 23.73 & 0.779 & 0.195 & 25.13 & 0.827 & 0.160 \\
SparseNeRF \cite{wang2023sparsenerf} & 19.86 & 0.624 & 0.328 & - & - & - & - & - & - \\
3DGS \cite{kerbl20233d} & 19.22 & 0.649 & 0.229 & 23.80 & 0.814 & 0.125 & 25.44 & 0.860 & 0.096 \\
DNGaussian \cite{li2024dngaussian} & 19.12 & 0.591 & 0.294 & 22.18 & 0.755 & 0.198 & 23.17 & 0.788 & 0.180 \\
FSGS \cite{zhu2023fsgs} & 20.43 & 0.682 & 0.248 & 24.09 & 0.823 & 0.145 & 25.31 & 0.860 & 0.122 \\
CoR-GS \cite{zhang2024corgssparseview3dgaussian} & 20.45 & 0.712 & \cellcolor{yellow!50}0.196 & 24.49 & 0.837 & \cellcolor{yellow!50}0.115 & 26.06 & 0.874 & 0.089 \\
DropGaussian \cite{dropgaussian} & \cellcolor{yellow!50}20.76 & \cellcolor{yellow!50}0.713 & 0.200 & \cellcolor{yellow!50}24.74 & \cellcolor{yellow!50}0.837 & 0.117 & \cellcolor{yellow!50}26.21 & \cellcolor{yellow!50}0.874 & \cellcolor{yellow!50}0.088 \\
CoMapGS \cite{comapgs} & \cellcolor{orange!50}21.11 & \cellcolor{orange!50}0.747 & \cellcolor{orange!50}0.182 & \cellcolor{red!50}25.20 & \cellcolor{red!50}0.854 & \cellcolor{orange!50}0.108 & \cellcolor{red!50}26.73 & \cellcolor{red!50}0.886 & \cellcolor{red!50}0.082 \\
\textbf{TWINGS (Ours)} & \cellcolor{red!50}21.49 & \cellcolor{red!50}0.754 & \cellcolor{red!50}0.167 & \cellcolor{orange!50}24.97 & \cellcolor{orange!50}0.844 & \cellcolor{red!50}0.106 & \cellcolor{orange!50}26.23 & \cellcolor{orange!50}0.876 & \cellcolor{orange!50}0.085 \\ \hline
\end{tabular}
}
\end{table}
}

\noindent{\textbf{Datasets.}} We evaluate our method on three representative benchmarks: LLFF~\cite{llff}, DTU~\cite{DTU}, and Mip-NeRF360~\cite{mipnerf360}. Following previous work~\cite{zhang2024corgssparseview3dgaussian,li2024dngaussian,dropgaussian}, we adopt the same splits with 3 training views for LLFF and DTU and 12 training views for Mip-NeRF360, and downsample the images to ×8, ×4, and ×4, respectively. 

\noindent{\textbf{Evaluation metrics.}} We evaluate novel view synthesis quality using peak signal-to-noise ratio (PSNR) for pixel-wise fidelity, SSIM~\cite{wang2004image} for structural similarity, and LPIPS~\cite{zhang2018unreasonable} for human-perceived visual differences via deep features, which have been generally adopted in sparse-view scenarios. We also compute the AVGE~\cite{muller2022instant}, which is the geometric mean of PSNR, SSIM, and LPIPS.


\noindent{\textbf{Baselines.}} We compare our method with sparse novel view synthesis approaches such as Mip-NeRF~\cite{barron2021mip}, DietNeRF~\cite{jain2021putting}, RegNeRF~\cite{niemeyer2022regnerf}, FreeNeRF~\cite{yang2023freenerf}, and SparseNeRF~\cite{wang2023sparsenerf}, as well as 3DGS-based methods including 3DGS~\cite{kerbl20233d}, FSGS~\cite{zhu2023fsgs}, CoR-GS~\cite{zhang2024corgssparseview3dgaussian}, DNGaussian~\cite{li2024dngaussian}, DropGaussian~\cite{dropgaussian}, and CoMapGS~\cite{comapgs}. Following previous work \cite{zhu2023fsgs, zhang2024corgssparseview3dgaussian}, we utilize a fused stereo point cloud from sparse views as the initial point cloud. For CoMapGS, we report the rendered results provided by the authors.

\noindent{\textbf{Implementation Details.}} Our 3DGS optimization incorporates the structural regularization strategy of DropGaussian \cite{dropgaussian}. Our total loss function is defined as:
\begin{equation}
\mathcal{L} = \mathcal{L}_1(\hat{I}, I) + \lambda_1 \mathcal{L}_{D\text{-}SSIM}(\hat{I}, I) + \lambda_2 \mathcal{L}_{D}(\hat{D}, D_{est}),
\end{equation}

\noindent{where} $L_1$ and $L_{D\text{-}SSIM}$ represent the photometric loss term between rendered image $\hat{I}$ and ground truth image $I$~\cite{dropgaussian}. $\mathcal{L}_{D}$ is depth loss between the rendered depth $\hat{D}$ and the estimated depth $D_{est}$~\cite{zhu2023fsgs, depthpro}, $\lambda_1$ and $\lambda_2$ are set to 0.2 and 0.01, respectively. 

\subsection{Comparison with State-of-the-Art Methods}
\noindent\textbf{DTU.} Quantitative comparisons on the DTU dataset are reported in Table~\ref{table:DTU}. Across the 3-/6-/9-view settings, TWINGS attains the highest PSNR and SSIM and the lowest LPIPS and AVGE, with the most pronounced gains in the 3-view setting, supporting the effectiveness of our dense, well-calibrated initialization. As illustrated in Fig.~\ref{fig:dtu_fig}, CoR-GS, DNGaussian, and DropGaussian fail to reconstruct the window framing of the building, misplace lettering relative to the geometry, and fail to recover black pupils. By contrast, our method reconstructs the window framing of the building and is particularly strong at preserving fine details, including geometrically accurate lettering and black pupils that competing methods fail to capture. 



\noindent\textbf{Mip-NeRF360.} In the 12-view setting, TWINGS achieves the highest PSNR and SSIM, as reported in Table~\ref{table:mipnerf360}. As illustrated in Fig.~\ref{fig:mipnerf_fig}, on the top row, CoR-GS, DropGaussian, and CoMapGS fail to reconstruct the white pillar and its cast shadow, whereas TWINGS faithfully recovers both. On the bottom row, CoR-GS mislocalizes the ball, DropGaussian renders it overly transparent, and CoMapGS reconstructs a sphere with a deformed top. All three methods over-smooth the grass, whereas TWINGS preserves the spherical shape of the ball and recovers fine structures in the grass.

\noindent\textbf{LLFF.} In the challenging 3-view setting, TWINGS outperforms competing methods, as reported in Table~\ref{table:llff}. As illustrated in Fig.~\ref{fig:llff_fig}, CoR-GS and DropGaussian fail to reconstruct the ceiling sprinkler, while CoMapGS renders it overly translucent. All three baselines introduce streaking artifacts on the small signboard and miss the intricate black-dot patterns on the white marble. TWINGS faithfully recovers these details, indicating that our geometry-calibrated dense initialization reduces over-smoothing during 3DGS optimization and preserves high-frequency detail under sparse supervision.



\subsection{Ablation Study}
\label{sec:ablation_study}

\noindent{\textbf{Impact of multi-view correspondences.}} We evaluate our multi-view correspondences by comparing them with a variant that generates only pairwise correspondences among the training views. As shown in Table~\ref{table:multiview_corres_ablation}, our multi-view correspondences yield globally coherent and accurate 3D points, which in turn improves subsequent TPS deformation and the quality of novel view synthesis.

{
\setlength{\floatsep}{4pt}        
\setlength{\textfloatsep}{6pt}    
\setlength{\intextsep}{6pt}       

\begin{table}[!htbp]
\centering
\caption{Ablation of pairwise and multi-view correspondences. Multi-view correspondences consistently improve novel view synthesis quality.}
\vspace{-5pt}
\label{table:multiview_corres_ablation}
\resizebox{\columnwidth}{!}{
\begin{tabular}{@{}c|cc|cc|cc@{}}
\toprule
\multirow{2}{*}{\textbf{Method}} & \multicolumn{2}{c|}{\textbf{\begin{tabular}[c]{@{}c@{}}DTU \\ (3-view)\end{tabular}}} & \multicolumn{2}{c|}{\textbf{\begin{tabular}[c]{@{}c@{}}LLFF \\ (3-view)\end{tabular}}} & \multicolumn{2}{c}{\textbf{\begin{tabular}[c]{@{}c@{}}MipNeRF-360 \\ (12-view)\end{tabular}}} \\ \cmidrule(l){2-7} 
 & \textbf{PSNR ↑} & \textbf{SSIM ↑} & \textbf{PSNR ↑} & \textbf{SSIM ↑} & \textbf{PSNR ↑} & \textbf{SSIM ↑} \\ \midrule
pairwise & 21.32 & 0.875 & 21.29 & 0.743 & 19.99 & 0.607 \\
multi-view & \textbf{21.52} & \textbf{0.880} & \textbf{21.49} & \textbf{0.754} & \textbf{20.35} & \textbf{0.618} \\ \bottomrule
\end{tabular}
}
\end{table}

\begin{table}[!htbp]
\centering
\caption{Impact of the TWINGS-Init on the sparse novel view synthesis (DTU 3-view). Quantitative comparison of GS baselines with and without TWINGS-Init.}
\vspace{-5pt}
\label{table:dtu_ablation} 
\resizebox{0.95\columnwidth}{!}{
\begin{tabular}{@{}c|cccc@{}}
\toprule
\textbf{Method} & \textbf{PSNR↑} & \textbf{SSIM↑} & \textbf{LPIPS↓} & \textbf{AVGE↓} \\ \midrule
\multicolumn{1}{c|}{3DGS} & \multicolumn{1}{c|}{17.65} & \multicolumn{1}{c|}{0.816} & \multicolumn{1}{c|}{0.146} & 0.108 \\
\multicolumn{1}{c|}{3DGS \textbf{+ TWINGS-Init}} & \multicolumn{1}{c|}{\textbf{20.21}} & \multicolumn{1}{c|}{\textbf{0.856}} & \multicolumn{1}{c|}{\textbf{0.120}} & \textbf{0.080} \\ \midrule
\multicolumn{1}{c|}{FSGS} & \multicolumn{1}{c|}{17.24} & \multicolumn{1}{c|}{0.824} & \multicolumn{1}{c|}{0.158} & 0.115 \\
\multicolumn{1}{c|}{FSGS \textbf{+ TWINGS-Init}} & \multicolumn{1}{c|}{\textbf{20.42}} & \multicolumn{1}{c|}{\textbf{0.866}} & \multicolumn{1}{c|}{\textbf{0.126}} & \textbf{0.079} \\ \midrule
\multicolumn{1}{c|}{CoR-GS} & \multicolumn{1}{c|}{19.21} & \multicolumn{1}{c|}{0.853} & \multicolumn{1}{c|}{0.119} & 0.087 \\
\multicolumn{1}{c|}{CoR-GS \textbf{+ TWINGS-Init}} & \multicolumn{1}{c|}{\textbf{21.27}} & \multicolumn{1}{c|}{\textbf{0.880}} & \multicolumn{1}{c|}{\textbf{0.104}} & \textbf{0.070} 
\\ \bottomrule
\end{tabular}
}
\end{table}
}

\begin{figure}
    \centering
    \includegraphics[width=\columnwidth]{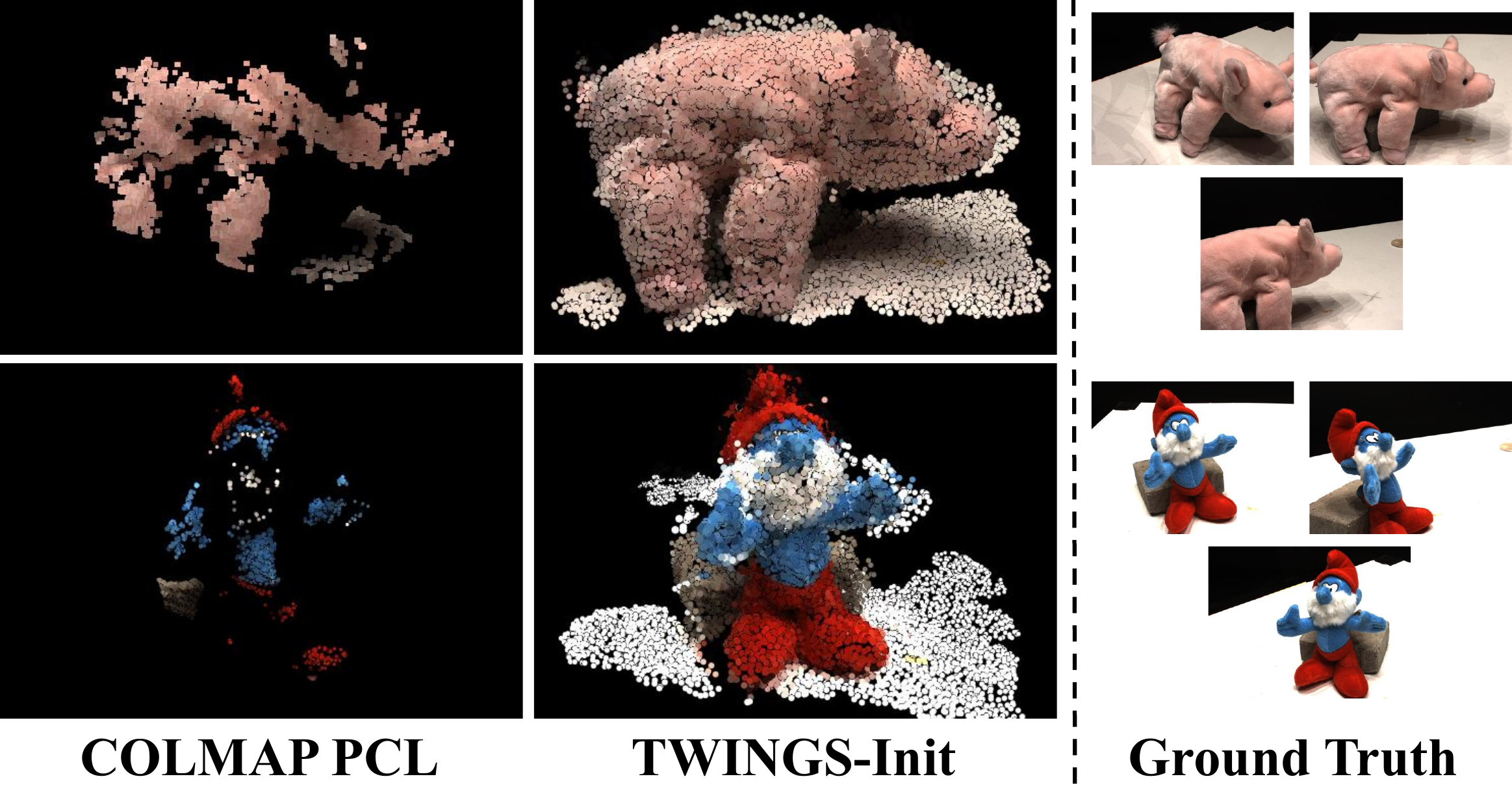}
    \vspace{-20pt}
    \caption{Visualization of reconstructed 3D point clouds (DTU 3-view). TWINGS-Init is substantially denser than the COLMAP PCL and accurately captures the scene geometry.}
    \label{fig:dtu_ablation}
\end{figure}

\begin{figure}[!htbp]
    \centering
    \includegraphics[width=0.95\columnwidth]{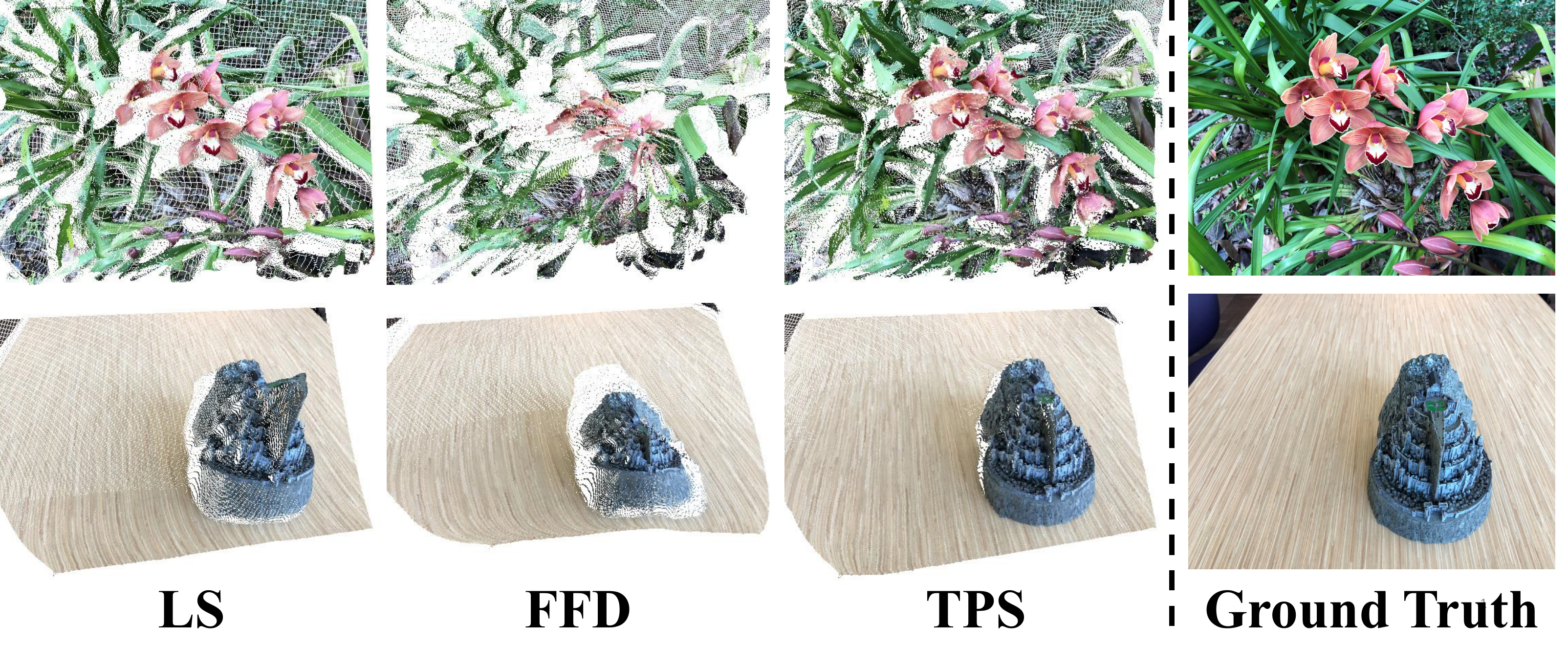}
    \vspace{-5pt}
    \caption{Visualization of deformation methods on the LLFF dataset. The rightmost column shows the ground truth training view to which deformation is applied, and the other columns visualize the corresponding deformed results from viewpoints that highlight geometric details.}
\label{fig:deformation_comparison}
\end{figure}

\noindent{\textbf{Impact of our Initialization Module.}} We ablate our plug-and-play initialization module, termed TWINGS-Init, on the DTU dataset in the 3-view setting to verify its effectiveness. Examples of the point cloud generated by TWINGS-Init are visualized in Fig.~\ref{fig:dtu_ablation}. As shown in Table~\ref{table:dtu_ablation}, TWINGS-Init demonstrates significant performance gains across all baselines. For instance, integrating TWINGS-Init with the vanilla 3DGS model boosts the PSNR by a remarkable +2.56 dB (17.65 $\rightarrow$ 20.21). Notably, this enhancement is substantial enough to make the vanilla 3DGS competitive with previous state-of-the-art methods. Furthermore, when applied to a stronger model like CoR-GS, TWINGS-Init improves the PSNR by +1.96 dB (19.21 $\rightarrow$ 21.17), surpassing the existing state of the art. This supports that our highly accurate and dense reconstructed 3D points stabilize the 3DGS optimization process and prevent overfitting to training views, improving novel view synthesis quality. Importantly, TWINGS-Init completes in 12.45 s and is seamlessly integrated into 3DGS methods with only a minimal additional time cost. Please refer to the supplementary material for details.

\noindent{\textbf{Impact of Deformation Methods.}} We compare deformation strategies for geometric alignment: LS, FFD, and TPS. As illustrated in Fig.~\ref{fig:deformation_comparison}, LS often produces unstable local geometry near the image center. FFD often introduces distortions that fail to preserve scene structures. In contrast, TPS models non-rigid variations effectively and yields geometrically accurate alignments. For all three variants, we apply CBPS after deformation and initialize our 3DGS with the sampled points. Table~\ref{table:deformation_ablation} shows that TPS achieves the best rendered results across all metrics.

{%
\setlength{\floatsep}{4pt}        
\setlength{\textfloatsep}{6pt}    
\setlength{\intextsep}{6pt}       

\begin{table}[!t]
\centering
\caption{Ablation on deformation methods on the LLFF dataset (3-view). Comparison of novel view synthesis results for TPS, FFD, and LS. TPS consistently outperforms the other methods across all metrics.}
\vspace{-5.0pt}
\label{table:deformation_ablation}
\scriptsize
\resizebox{0.90\columnwidth}{!}{%
\begin{tabular}{c|c|c|c}
\toprule
\textbf{Method} & \textbf{PSNR ↑} & \textbf{SSIM ↑} & \textbf{LPIPS ↓} \\ \midrule
Linear Scaling   & {\scriptsize 20.90} & {\scriptsize 0.725} & {\scriptsize 0.189} \\
Free Form Deformation   & {\scriptsize 20.96} & {\scriptsize 0.727} & {\scriptsize 0.187} \\
TPS             & {\scriptsize \textbf{21.49}} & {\scriptsize \textbf{0.754}} & {\scriptsize \textbf{0.167}} \\
\bottomrule
\end{tabular}
}
\end{table}


\begin{table}[!t]
\centering
\caption{Ablation on the sampling radius $r$ on the LLFF dataset with 3, 6, 9 training views. The radius is set to a fraction of $S$.}
\vspace{-5.0pt}
\label{table:cbps_ablation}
\renewcommand{\arraystretch}{1.2}
\scriptsize
\resizebox{\columnwidth}{!}{%
\begin{tabular}{@{}c|cc|cc|cc@{}}
\toprule
\multirow{2}{*}{\textbf{Sampling Radius ($r$)}} & \multicolumn{2}{c|}{\textbf{3-view}} & \multicolumn{2}{c|}{\textbf{6-view}} & \multicolumn{2}{c}{\textbf{9-view}} \\ \cmidrule(l){2-7} 
 & \textbf{PSNR ↑} & \textbf{SSIM ↑} & \textbf{PSNR ↑} & \textbf{SSIM ↑} & \textbf{PSNR ↑} & \textbf{SSIM ↑} \\ \midrule
$S \cdot 1/8$ & \scriptsize 21.49 & \scriptsize 0.754 & \scriptsize 24.97 & \scriptsize 0.844 & \scriptsize 26.12 & \scriptsize 0.875 \\
$S \cdot 1/16$ & \scriptsize 21.22 & \scriptsize 0.746 & \scriptsize 24.89 & \scriptsize 0.843 & \scriptsize 26.23 & \scriptsize 0.876 \\
$S \cdot 1/32$ & \scriptsize 21.23 & \scriptsize 0.739 & \scriptsize 24.93 & \scriptsize 0.843 & \scriptsize 26.16 & \scriptsize 0.875 \\
$S \cdot 1/64$ & \scriptsize 21.07 & \scriptsize 0.734 & \scriptsize 24.92 & \scriptsize 0.842 & \scriptsize 26.07 & \scriptsize 0.875 \\ \bottomrule
\end{tabular}
}
\end{table}
}

\noindent{\textbf{Impact of Sampling Radius.}} As shown in Table~\ref{table:cbps_ablation}, the optimal sampling radius $r$ depends on the geometric configuration of the input views. In the 3-view setting, where the COLMAP PCL is extremely sparse, a larger radius is required to ensure sufficient point density. In contrast, with 9 views, the COLMAP PCL becomes denser and more accurate. In this regime, a larger radius can be suboptimal, as 3DGS favors the quality of the initial points over their sheer quantity. Please refer to the supplementary material for qualitative comparisons.



\label{sec:experiments}

\section{Conclusion}

In this paper, we present TWINGS, a framework that tackles point sparsity in sparse-view 3DGS by converting monocular depth into a geometry-consistent point set via TPS-based warp alignment to triangulated geometry. TWINGS triangulates multi-view correspondences to build control points, warps backprojected depth into calibrated backprojected points, and samples them near reliable controls to produce a dense, geometrically accurate initialization that markedly stabilizes 3DGS training under extremely sparse-view scenarios.

\noindent{\textbf{Limitations.}} Our method targets point sparsity in extremely sparse-view scenarios where the initial COLMAP PCL poorly captures geometry. The benefit tapers off as views increase or in texture-rich scenes where SfM reliably recovers geometry. As future work, the framework can be extended to 3-view surface reconstruction for real-time AR/VR, content creation~\cite{GHG, THuman}, leveraging our fast and accurate initialization to preserve fine facial details.

\noindent \textbf{Acknowledgements.} This work was supported by the National Research Foundation of Korea(NRF) grant funded by the Korea government(MSIT) (RS-2025-00573117, RS-2025-25442295, RS-2025-16070382). The work was supported by the Korea Institute of Science and Technology (KIST) Institutional Program under Grant 26E0170, 26E0171. This research was supported by Artificial Intelligence Graduate School Program at Yonsei University (RS-2020-II201361).



\label{sec:conclusion}

{
    \small
    \bibliographystyle{ieeenat_fullname}
    \bibliography{main}
}


\clearpage
\clearpage
\setcounter{page}{1}
\maketitlesupplementary

\setcounter{section}{0}
\renewcommand{\thesection}{\Alph{section}}

\section{Time Complexity}
\label{sec:time_complexity}

All preprocessing timings were measured on a local workstation equipped with an NVIDIA GeForce RTX 3060 GPU and an Intel Core i5-12400 CPU, while 3DGS training was performed on a server equipped with an NVIDIA RTX 6000 Ada GPU and an AMD EPYC 7763 CPU.
Table~\ref{tab:time_complexity} summarizes the runtime of each stage in the preprocessing pipeline.
The total preprocessing time is 12.45 s for DTU (3-view), 17.74 s for LLFF (3-view), and 94.50 s for Mip-NeRF360 (12-view).
In comparison, our 3DGS training takes 12.75 min on DTU (3-view), 12.77 min on LLFF (3-view), and 14.40 min on Mip-NeRF360 (12-view).
The preprocessing time increases roughly linearly with the number of input views, primarily due to multi-view correspondence establishment and TPS deformation.
Even with 12 views, preprocessing completes in under 1.6 min, which remains negligible compared to the 3DGS training time.
For the challenging 3-view settings, preprocessing accounts for only about 1–2\% of the total training time, demonstrating the efficiency and practicality of our pipeline for real-world applications.

\begin{table}[!htbp]
\centering
\caption{Runtime of each stage in the preprocessing pipeline across datasets and various view settings.}
\label{tab:time_complexity}
\scriptsize
\resizebox{0.95\columnwidth}{!}{%
\begin{tabular}{@{}l|ccc@{}}
\toprule
\multirow{2}{*}{\textbf{Stage}} & \multicolumn{3}{c}{\textbf{Processing Time (sec)}} \\ \cmidrule(l){2-4} 
 & \textbf{\begin{tabular}[c]{@{}c@{}}DTU\\ (3-view)\end{tabular}} & \textbf{\begin{tabular}[c]{@{}c@{}}LLFF\\ (3-view)\end{tabular}} & \textbf{\begin{tabular}[c]{@{}c@{}}Mip-NeRF360\\ (12-view)\end{tabular}} \\ \midrule
Multi-view correspondences & 4.78 & 5.96 & 41.69 \\
Multi-view triangulation & 0.97 & 1.31 & 5.28 \\
TPS deformation & 6.65 & 10.43 & 47.36 \\
CBPS & 0.05 & 0.04 & 0.17 \\ \midrule
\textbf{Total} & \textbf{12.45} & \textbf{17.74} & \textbf{94.50} \\ \bottomrule
\end{tabular}
}
\end{table}

We report the training time of GS-variants with 3-/6-/9-view on the DTU dataset, using the same PCL generated by TWINGS-Init. As reported in Table~\ref{table:efficiency_analysis}, our method remains highly efficient as the number of views increases, showing a favorable scalability. 

\begin{table}[h]
\centering
\caption{Training time of GS-variants on the DTU dataset under different training view settings.}
\vspace{-2mm}
\label{table:efficiency_analysis}
\resizebox{0.70\columnwidth}{!}{
\begin{tabular}{@{}l|ccc@{}}
\toprule
Training Time (min) & 3-view & 6-view & 9-view \\ \midrule
FSGS & 35.48  & 35.86 & 32.01 \\
CoR-GS & 9.57  & 10.45 & 11.35 \\ 
TWINGS (Ours) & 12.74 & 12.81 & 13.22 \\ \bottomrule
\end{tabular}
}
\end{table}

\begin{figure}[!htbp]
    \centering
    \includegraphics[width=\columnwidth]{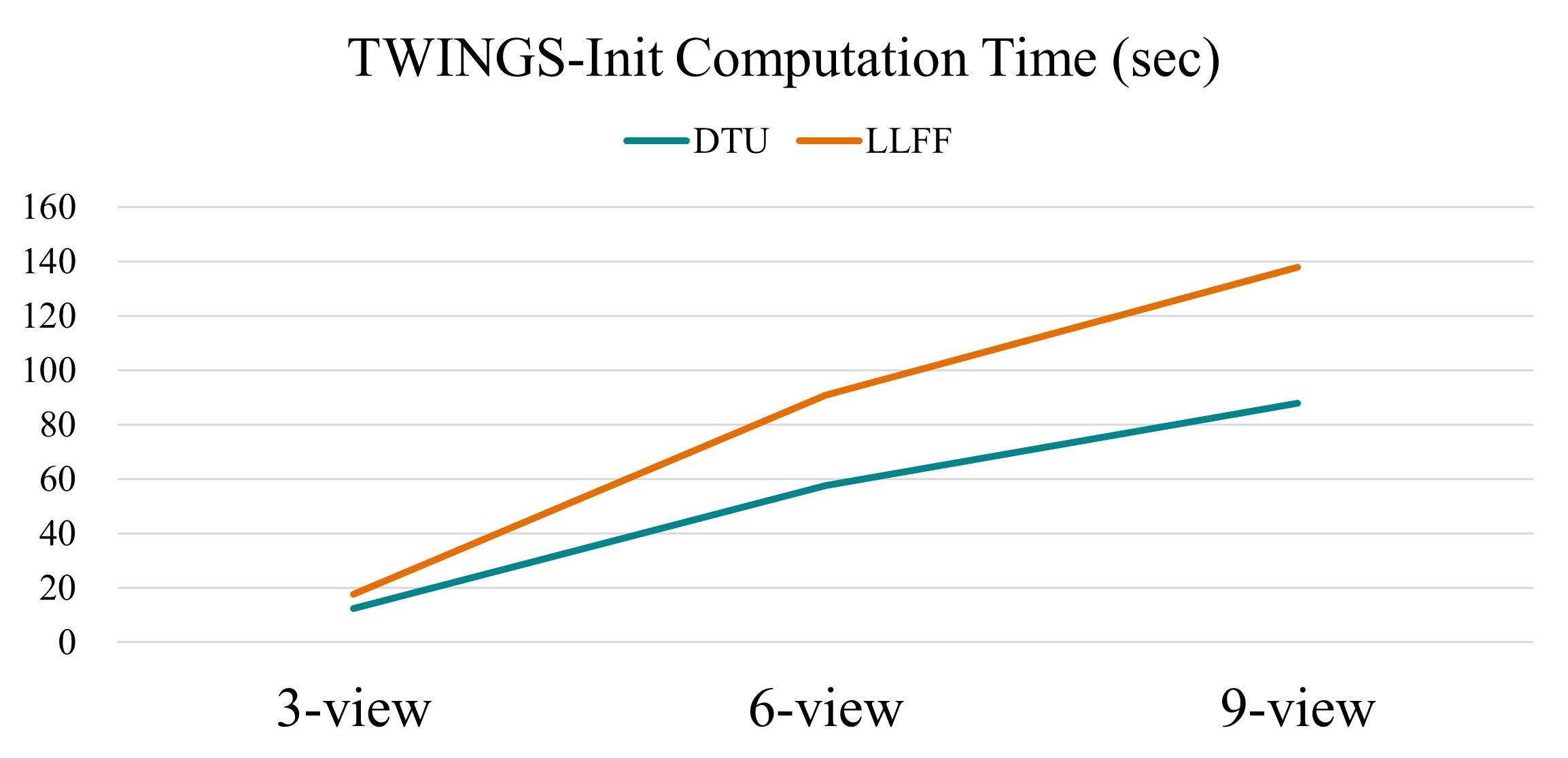}
    \caption{Computation time of TWINGS-Init on the LLFF and DTU datasets. The runtime grows approximately linearly with the number of input views, demonstrating the favorable scalability and efficiency of TWINGS-Init.}
    \label{fig:supple_twings-init_time}
\end{figure}

We report the computation time of TWINGS-Init across 3-/6-/9-view on the LLFF and DTU datasets. As illustrated in Fig.~\ref{fig:supple_twings-init_time}, the runtime scales approximately linearly with the number of input views, indicating that TWINGS-Init remains computationally efficient as the view count increases.

\section{Comparison with modern dense matchers}
\label{sec:supple_modern_dense_matchers}
Our contribution is to demonstrate how geometrically accurate initialization affects the optimization dynamics of 3DGS under sparse-view supervision. We observe that points triangulated from dense-matcher-derived matches tend to concentrate in foreground regions with high-confidence, while background regions remain sparsely covered. To address this imbalance, we use dense depth via LS or FFD to cover background regions, which results in suboptimal 3DGS performance as shown in Tab.~6 of the manuscript. This observation motivates our TPS with CBPS, which preserves reliable foreground geometry while additionally enabling the sampling of geometrically consistent points in background regions, even when only a sparse number of background matches are available. We find that such balanced coverage of both foreground and background regions is particularly effective in constraining Gaussian optimization in sparse-view scenarios.

\begin{table}[h]
\centering
\caption{Comparison with modern dense matchers.}
\vspace{-2mm}
\label{table:dense_matchers}
\resizebox{0.95\columnwidth}{!}{
\begin{tabular}{@{}l|cc|cc|cc@{}}
\toprule
\multirow{2}{*}{\textbf{Initial PCL}} & \multicolumn{2}{c|}{\begin{tabular}[c]{@{}c@{}}\textbf{DTU} \\ \textbf{(3-view)}\end{tabular}} & \multicolumn{2}{c|}{\begin{tabular}[c]{@{}c@{}}\textbf{LLFF} \\ \textbf{(3-view)}\end{tabular}} & \multicolumn{2}{c}{\begin{tabular}[c]{@{}c@{}}\textbf{MipNeRF-360}\\ \textbf{(12-view)}\end{tabular}} \\ \cmidrule(l){2-7} 
 & PSNR↑ & SSIM↑ & PSNR↑ & SSIM↑ & PSNR↑ & SSIM↑ \\ \midrule
MASt3R & 21.05 & 0.853 & 20.26 & 0.702 & 19.78 & 0.573 \\
\textbf{TWINGS-Init} & \textbf{21.52} & \textbf{0.880} & \textbf{21.49} & \textbf{0.754} & \textbf{20.35} & \textbf{0.618} \\ \bottomrule
\end{tabular}
}
\end{table}
\vspace{-2mm}

We compare TWINGS-Init with point clouds triangulated from MASt3R~\cite{mast3r} matches on three benchmark datasets. We evaluate TWINGS with an identical training pipeline, varying only the initial PCL used for Gaussian initialization. As reported in Table~\ref{table:dense_matchers}, using TWINGS-Init consistently yields superior novel view synthesis results compared to modern dense matchers in sparse-view scenarios.

\section{Different matching algorithms}
\label{sec:supple_matching}
\begin{figure}[!htbp]
    \centering
    \includegraphics[width=\columnwidth]{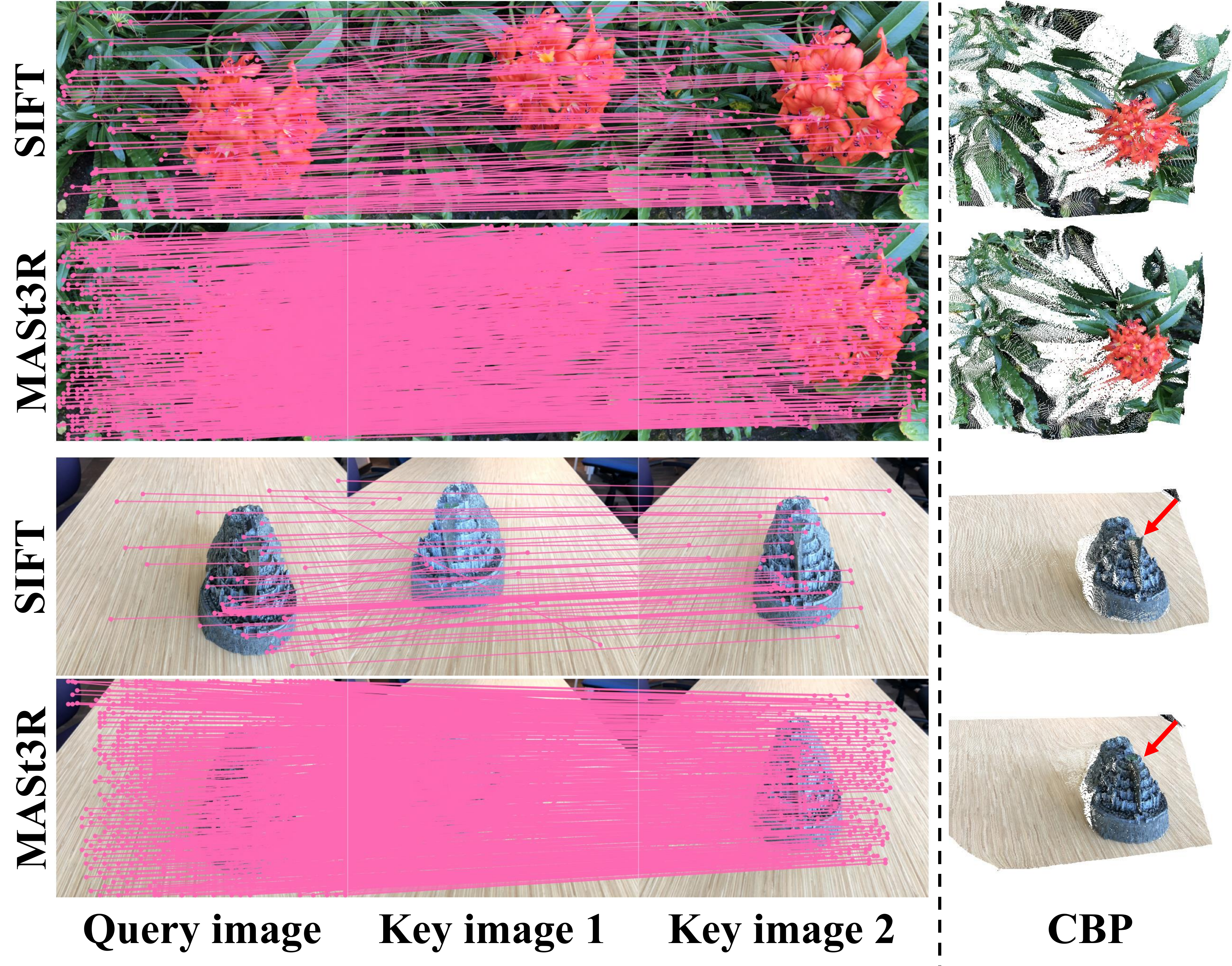}
    \caption{Visualization of CBP with different matching algorithms. The leftmost column shows the query image, the second and third columns show the key images, and the rightmost column shows the CBP of the query view to which TPS deformation is applied, shown from viewpoints that highlight geometric details.}
    \label{fig:supple_matching}
\end{figure}

We ablate the correspondence module in our pipeline by replacing MASt3R~\cite{mast3r} with SIFT~\cite{Schonberger_2016_CVPR} as used in COLMAP to assess how sensitive the pipeline is to the quality of feature correspondences. As illustrated in Fig.~\ref{fig:supple_matching}, the SIFT replacement yields sparser and less reliable matches. When a sufficient number of reliable correspondences are established, our method still reconstructs a coarse geometry (CBP). However, when the matches are few or unreliable, the geometry tends to be slightly distorted. In \textit{Flower} scene, the petal shapes appear elongated, and in \textit{Fortress} scene, the frontal region of the fortress is reconstructed with an unnaturally sharp protrusion compared to the ground truth images. 

\begin{table}[!htbp]
\centering
\caption{Ablation study on the correspondence module on the LLFF dataset (3-view). Improved correspondence quality in our pipeline results in better rendering performance.}
\label{table:supple_matching}
\scriptsize
\setlength{\tabcolsep}{8pt} 
\resizebox{0.85\columnwidth}{!}{
\begin{tabular}{c|ccc} 
\toprule
\multirow{2}{*}{\textbf{Image Matcher}} & \multicolumn{3}{c}{\textbf{LLFF (3-view)}} \\ \cmidrule(l){2-4} 
 & \textbf{PSNR ↑} & \textbf{SSIM ↑} & \textbf{LPIPS↓} \\ \midrule
SIFT & 21.01 & 0.732 & 0.187 \\
MASt3R & \textbf{21.49} & \textbf{0.754} & \textbf{0.167} \\ \bottomrule
\end{tabular}
}
\end{table}

In contrast, MASt3R produces denser and more robust correspondences across wide viewpoint changes, thereby enabling TPS model definition to be better established and yielding more accurate 3D geometry. The resulting gains in rendering quality are reported in Table~\ref{table:supple_matching}.

\section{Discussion on the LLFF dataset}
\label{sec:supple_covis}
To further analyze the diminishing effect of our method on the LLFF dataset with increasing views (Sec.~5.3, “Impact of Sampling Radius”), we introduce a multi-view score inspired by the covisibility map formulation of CoMapGS~\cite{comapgs}. Specifically, we compute the proportion of pixels whose covisibility count exceeds two (i.e., observed by at least three views), which measures the extent of reliable multi-view correspondences within the dataset. A higher multi-view score indicates that a greater portion of pixels are consistently observed across multiple views, thereby providing stronger geometric constraints during SfM.

\begin{table}[!htbp]
\centering
\caption{Multi-view scores on the benchmark datasets. As the number of views increases, the LLFF dataset exhibits a substantially higher multi-view scores, reflecting its forward-facing camera configuration and high inter-view overlap.}
\label{table:supple_covis}
\scriptsize
\resizebox{0.95\columnwidth}{!}{
\begin{tabular}{@{}c|ccccc@{}}
\toprule
\multirow{2}{*}{\textbf{Dataset}} & \multicolumn{5}{c}{\textbf{Multi-view scores}} \\ \cmidrule(l){2-6} 
 & \textbf{3-view} & \textbf{6-view} & \textbf{9-view} & \textbf{12-view} & \textbf{24-view} \\ \midrule
LLFF & 0.512 & \textbf{0.828} & \textbf{0.906} & - & - \\
DTU & 0.329 & 0.674 & 0.784 & - & - \\
MipNeRF-360 & - & - & - & 0.529 & 0.765 \\ \bottomrule
\end{tabular}
}
\end{table}

Unlike DTU and MipNeRF-360 datasets, where camera poses exhibit wider baselines or full 360-degree distributions, the LLFF dataset follows a forward-facing configuration in which cameras are concentrated along a limited arc in front of the scene. This geometric property significantly affects the covisibility pattern, as even modest increases in the number of training views yield a large boost in covisible regions, as adjacent views share a high degree of pixel overlap. Consequently, the multi-view scores of the LLFF dataset grows more rapidly compared to other datasets, as shown in Table~\ref{table:supple_covis}. When the view count increases in the LLFF dataset, the densified COLMAP point cloud becomes not only denser but also more reliable due to increased multi-view agreement. In this regime, the sampling radius $r$ in CBPS plays a lesser role, as the initial geometry already benefits from robust correspondences across views. In contrast, at very sparse settings (3-view), the lack of overlapping observations produces fewer covisible pixels, and thus a larger sampling radius is required to compensate for missing geometric cues. Our method particularly benefits from such low-covisibility conditions, providing a dense and accurate initialization that greatly stabilizes 3DGS optimization.

\begin{figure*}[!htbp]
    \centering
    \includegraphics[width=0.99\textwidth]{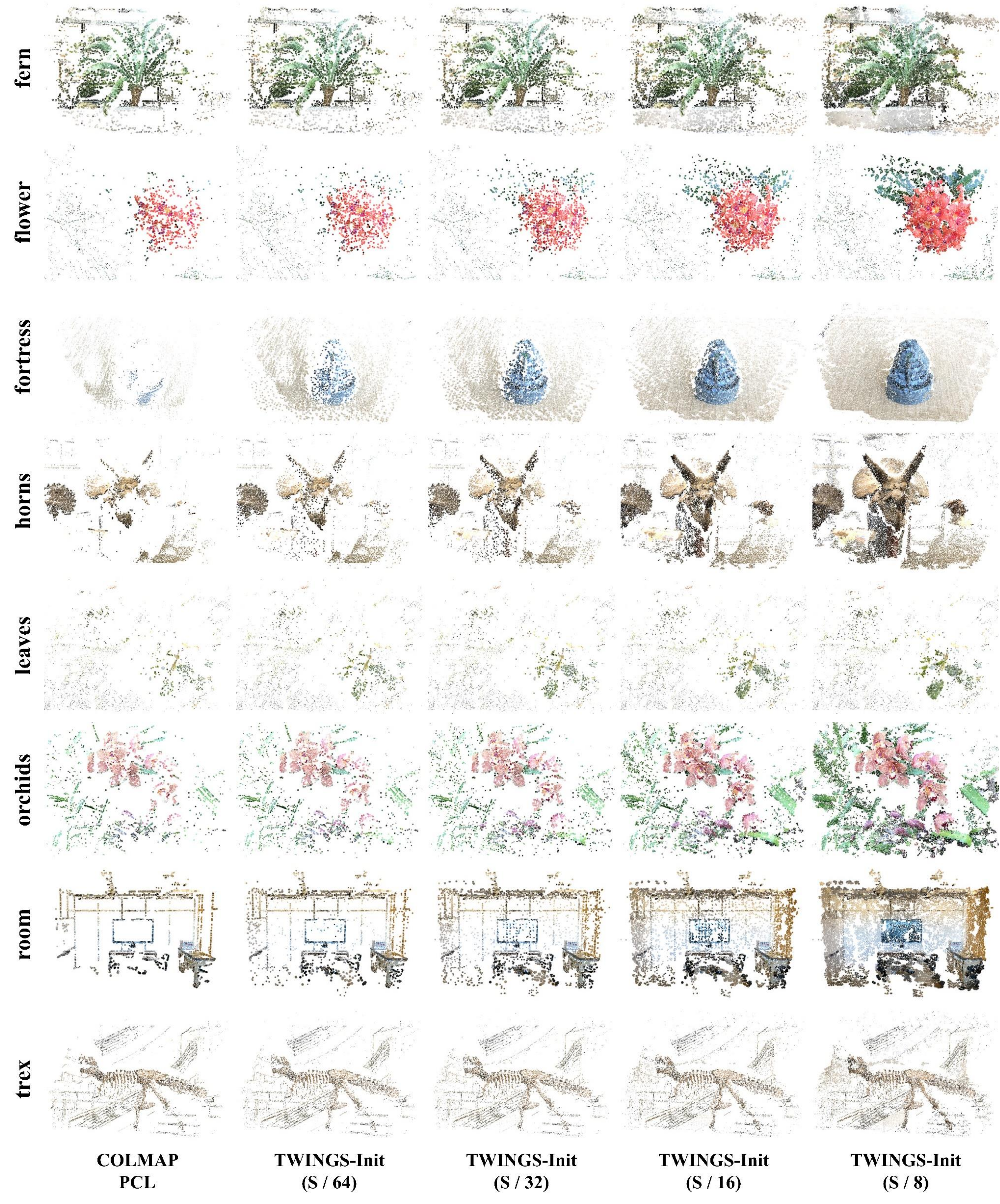}
    \caption{Point cloud comparison with varying CBPS sampling distances on the LLFF dataset. For each scene, given 3 training views, the left-most column shows the COLMAP PCL, while the subsequent columns demonstrate the results from TWINGS-Init using different sampling distances.}
    \label{fig:suppl_TWINGS}
\end{figure*}

\section{Visualization of TWINGS-Init}
\label{sec:supple_TWINGS}

We compare the COLMAP PCL and the point clouds from TWINGS-Init under different CBPS sampling distances. Results on the LLFF dataset with 3 training views are shown in Fig.~\ref{fig:suppl_TWINGS}. TWINGS-Init yields richer geometric detail and increased point concentration around scene structures. Reducing the sampling distance further concentrates samples near reliable control points, enhancing local fidelity and suppressing spurious samples.

\section{Additional Results}
\label{sec:additional_nvs_results_suppl}

\begin{figure}[!htbp]
    \includegraphics[width=0.96\columnwidth]{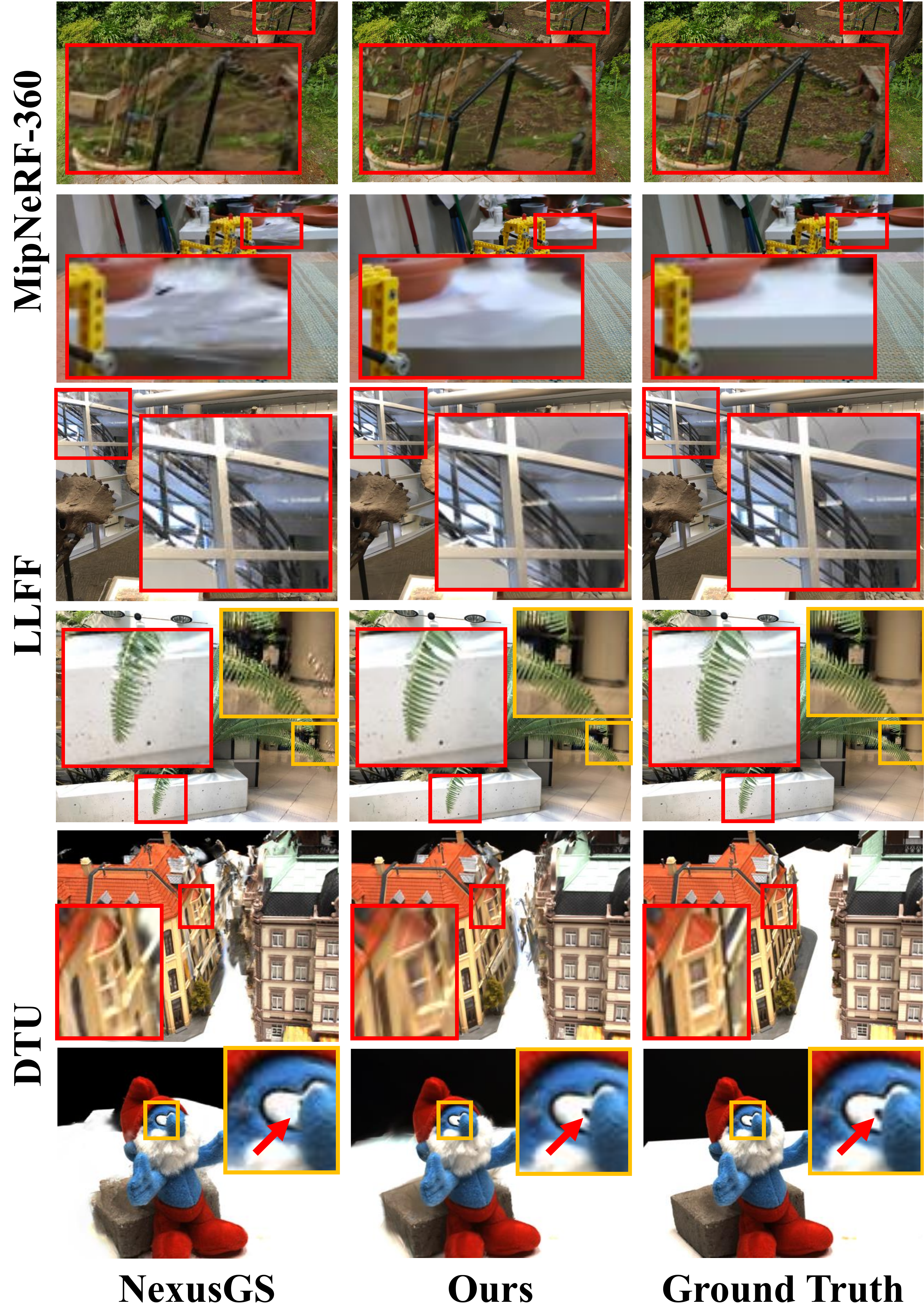}
    \caption{Qualitative comparison on the benchmark datasets. Novel view synthesis results rendered by NexusGS, our approach, and ground truth for comparison.}
    \vspace{-5mm}
    \label{fig:supple_nexus_gs}
\end{figure}

We provide additional rendering results. The examples on the DTU and LLFF datasets with 3 training views and the Mip-NeRF360 dataset with 12 training views are shown in Fig.~\ref{fig:supple_dtu_nvs}, Fig.~\ref{fig:supple_llff_nvs}, and Fig.~\ref{fig:supple_mipnerf360_nvs}, respectively. We additionally report comparisons against NexusGS~\cite{nexusgs} and Binocular3DGS~\cite{binocular3dgs} using their publicly available code under the same experimental setting. The metrics and rendered results are summarized in Table~\ref{table:supple_nexus_gs} and Fig.~\ref{fig:supple_nexus_gs}, where our method consistently achieves higher PSNR/SSIM and sharper, more faithful reconstructions.

\vspace{-2mm}

\begin{table}[!htbp]
\centering
\caption{Quantitative comparison of novel view synthesis results on the benchmark datasets.}
\label{table:supple_nexus_gs}
\scriptsize
\resizebox{\columnwidth}{!}{
\begin{tabular}{@{}c|cc|cc|cc@{}}
\toprule
\multirow{2}{*}{\textbf{Method}} & \multicolumn{2}{c|}{\textbf{\begin{tabular}[c]{@{}c@{}}DTU \\ (3-view)\end{tabular}}} & \multicolumn{2}{c|}{\textbf{\begin{tabular}[c]{@{}c@{}}LLFF \\ (3-view)\end{tabular}}} & \multicolumn{2}{c}{\textbf{\begin{tabular}[c]{@{}c@{}}MipNeRF-360 \\ (24-view)\end{tabular}}} \\ \cmidrule(l){2-7} 
 & \textbf{PSNR ↑} & \textbf{SSIM ↑} & \textbf{PSNR ↑} & \textbf{SSIM ↑} & \textbf{PSNR ↑} & \textbf{SSIM ↑} \\ \midrule
FSGS & 17.24 & 0.824 & 20.43 & 0.682 & 23.28 & 0.715 \\
CoR-GS & 19.21 & 0.853 & 20.45 & 0.712 & 23.39 & 0.727 \\
DropGaussian & 18.41 & 0.849 & 20.76 & 0.713 & 24.13 & 0.762 \\
NexusGS & 20.21 & 0.869 & 21.07 & 0.738 & 23.86 & 0.753 \\
Binocular3DGS & 20.71 & 0.862 & 21.44 & 0.751 & 22.26 & 0.700 \\
\textbf{TWINGS (Ours)} & \textbf{21.52} & \textbf{0.880} & \textbf{21.49} & \textbf{0.754} & \textbf{24.17} & \textbf{0.762} \\ \bottomrule
\end{tabular}
}
\end{table}

\begin{figure}[htbp!]
    \centering
    \includegraphics[width=\columnwidth]{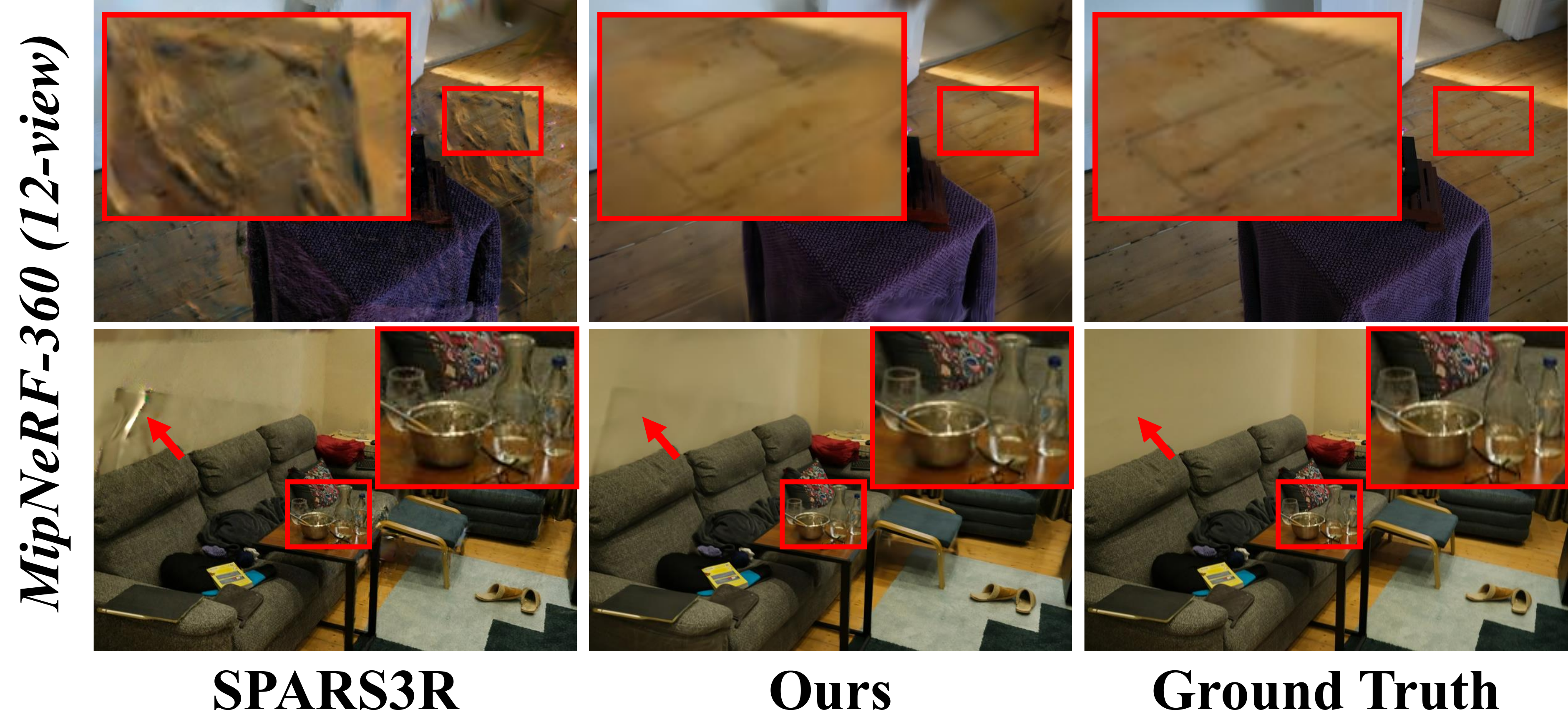}
    \vspace{-5mm}
    \caption{Qualitative comparison with SPARS3R.}
    \label{fig:spars3r}
\end{figure}

\begin{figure*}
    \centering
    \includegraphics[width=\textwidth]{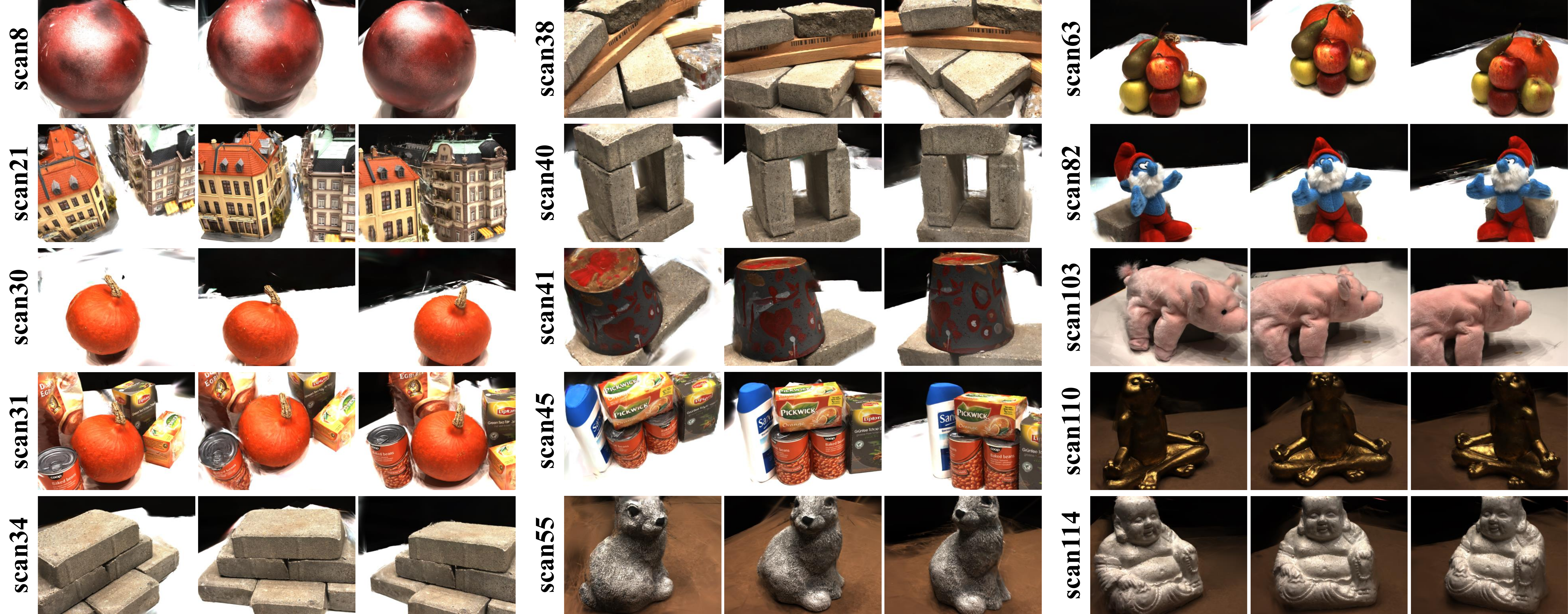}
    \caption{Examples of the rendered novel view results from TWINGS with 3 training views on the DTU dataset.}
    \label{fig:supple_dtu_nvs}
\end{figure*}

\begin{figure*}
    \centering
    \includegraphics[width=\textwidth]{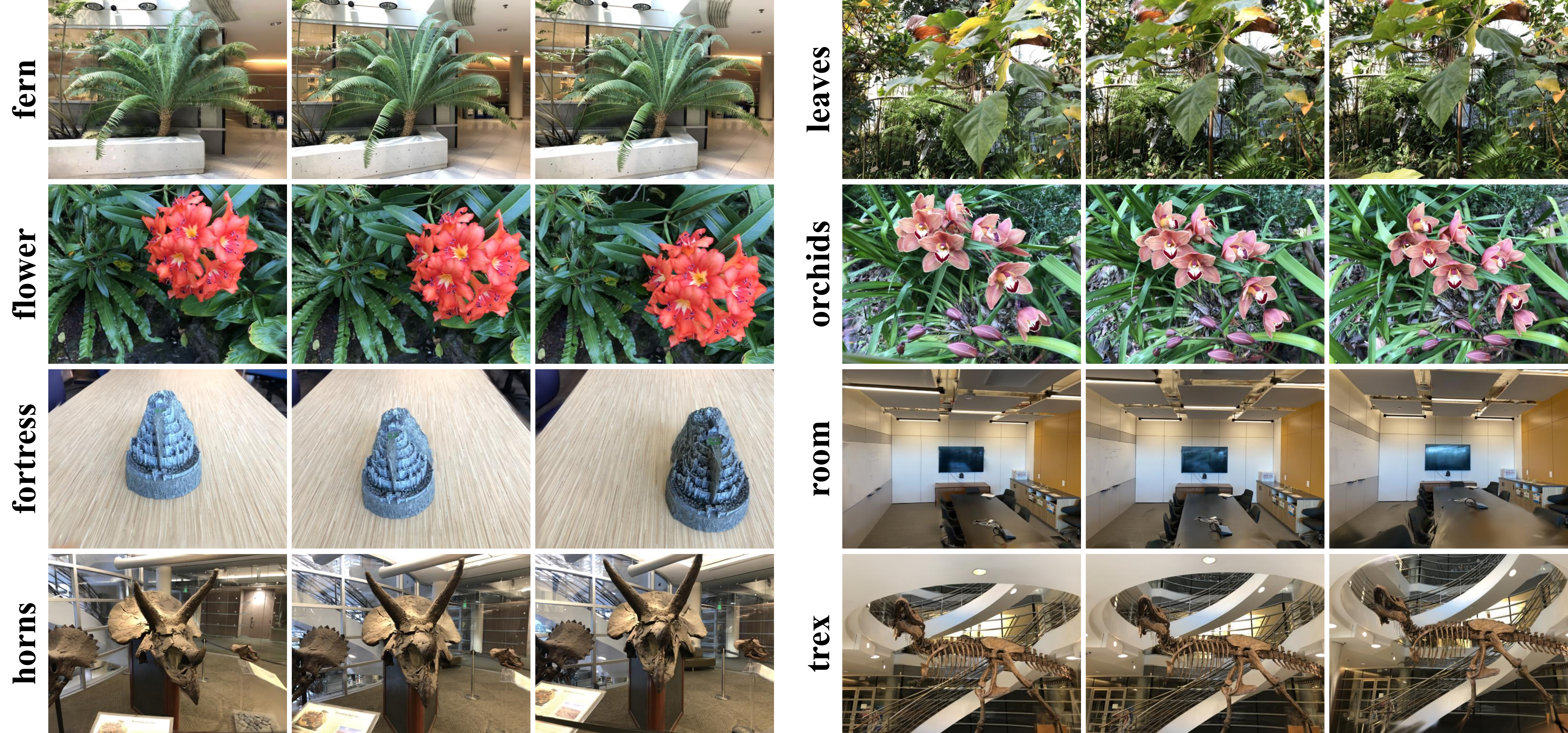}
    \caption{Examples of the rendered novel view results from TWINGS with 3 training views on the LLFF dataset.}
    \label{fig:supple_llff_nvs}
\end{figure*}

\begin{figure*}
    \centering
    \includegraphics[width=\textwidth]{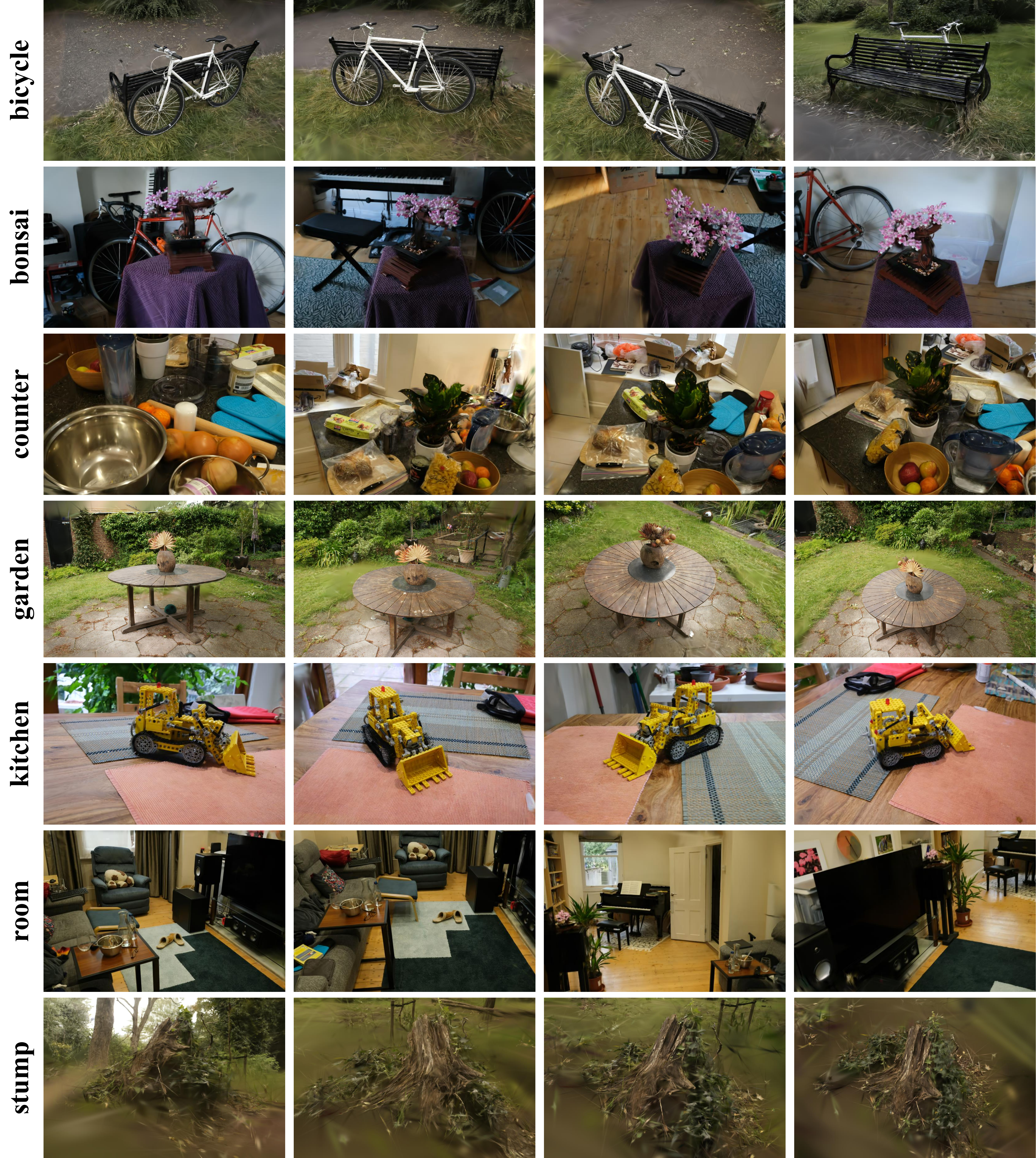}
    \caption{Examples of the rendered novel view results from TWINGS with 12 training views on the Mip-NeRF360 dataset.}
    \label{fig:supple_mipnerf360_nvs}
\end{figure*}

We additionally compare our method with recent sparse-view initialization methods. We compare our method with Dust-GS~\cite{dustgs} and SPARS3R~\cite{spars3r} on the same data split. As reported in Table~\ref{table:spars3r}, our method consistently achieves higher PSNR/SSIM on the MipNeRF-360 dataset. As illustrated in Fig.~\ref{fig:spars3r}, SPARS3R exhibits noticeable surface irregularities on planar regions such as walls and floors. We attribute this to residual geometric inconsistencies in its aligned PCL from MASt3R, leading to surface artifacts.

\begin{table}[h]
\centering
\renewcommand{\arraystretch}{1.0}
\caption{Comparisons with dense-initialization methods.}
\vspace{-2mm}
\label{table:spars3r}
\resizebox{0.95\columnwidth}{!}{
\begin{tabular}{@{}cccc@{}}
\toprule
\textbf{MipNeRF-360 (12-view)} & \textbf{Dust-GS} & \textbf{SPARS3R} & \textbf{TWINGS (Ours)} \\ \midrule
PSNR↑ / SSIM↑ & 18.75 / 0.552 & 19.29 / 0.578 & 20.35 / 0.618 \\ \bottomrule
\end{tabular}
}
\end{table}

\vspace{-2mm}
\vspace{-3mm}

\vspace{-2mm}
\section{Implementation Details}
\label{sec:implementation_details_suppl}
\noindent{\textbf{Thin Plate Splines.}} TPS is applied once to each training viewpoint, aligning backprojected points with reconstructed 3D points. This ensures that geometric information from all training viewpoints is incorporated, enhancing scene coverage.

\noindent{\textbf{Details of Point Sampling.}} To enhance the initialization of 3D Gaussian primitives, we combine the sparse 3D reconstruction produced by COLMAP with calibrated back-projected points (CBP). We refine this combined point set by removing points within a margin of 0.05 from the SfM reconstruction using a KD-tree to avoid overlap and applying radius-based clustering with a radius of 0.01 to reduce redundancy. For efficiency we randomly downsample the back-projected points to 30,000, a step that empirically causes no noticeable difference in the final geometric representation while improving computational efficiency.

\noindent\textbf{View Selection.}  
We select key images for each query view using a \(k\)-nearest-neighbor search in spherical coordinates.  
Each camera center in world coordinates is $\mathbf c_i=(x_i,y_i,z_i)$ and we define
$\alpha_i = \operatorname{atan2}(y_i, x_i)$ for azimuth,
$\varepsilon_i = \arcsin \bigl(z_i / r_i \bigr)$ for elevation,
and $r_i = \|\mathbf c_i\|_2$ for the distance from the world origin.  
For a query view $i$, the distance to every training view $j$ is as follows:
\begin{equation}
d_{ij} = \sqrt{(\alpha_i-\alpha_j)^2 + (\varepsilon_i-\varepsilon_j)^2 + (r_i-r_j)^2 } .
\end{equation}

We compute $d_{ij}$ to all training views and select the $k$ nearest neighbors as key images, with $k=2$ for the 3-view setting and $k=4$ for the 6-, 9-, 12-, and 24-view settings, giving a total of three or five images for multi-view triangulation.
This distance accounts for both viewing-direction differences and the radial distance of each camera center. Views that share a similar orientation but lie much farther or closer to the world origin than the query view, or that are at a similar radius but face a very different direction, result in larger $d_{ij}$ and are less likely to be selected. The method is particularly effective in sparse-view settings where each selected neighbor must provide strong image overlap and a well-conditioned baseline for reliable multi-view reconstruction.

\noindent{\textbf{Impact of View Selection Methods.}}
We perform an ablation on view selection strategies for multi-view triangulation on the Mip-NeRF360 dataset. As shown in Table~\ref{tab:triangulation_view_selection_mipnerf360}, nearest view selection yields consistently better performance than random selection. Nearest neighbors ensure sufficient overlap and moderate parallax, which provide denser and more reliable correspondences for triangulation. In contrast, random selection often introduces wide baselines or large angular differences, leading to fewer valid matches and noisier geometry that degrades subsequent 3DGS optimization. We experimentally find that nearest view selection produces more stable triangulation in sparse-view settings, and adopt this strategy for all experiments in this work.

\begin{table}[!htbp]
\centering
\caption{Ablation on view selection for multi-view triangulation on Mip-NeRF360 (12-view). Comparison of novel view synthesis results across different view selection strategies.}
\label{tab:triangulation_view_selection_mipnerf360}
\scriptsize
\resizebox{0.80\columnwidth}{!}{%
\begin{tabular}{c|c|c|c}
\toprule
\textbf{View Selection} & \textbf{PSNR$\uparrow$} & \textbf{SSIM$\uparrow$} & \textbf{LPIPS$\downarrow$} \\ \midrule
Random                  & 20.00                   & 0.605                   & 0.380                      \\
Nearest                 & \textbf{20.35}          & \textbf{0.617}          & \textbf{0.368}             \\ \bottomrule
\end{tabular}
}
\end{table}

\noindent{\textbf{Camera Poses.}} Following existing works \cite{niemeyer2022regnerf, wang2023sparsenerf}, we assume that all camera poses are known. In practice, for the DTU, LLFF, and Mip-NeRF360 datasets, we utilize the poses provided with the datasets.

\noindent{\textbf{Training.}} We implemented TWINGS using the Pytorch framework. During optimization, we densify the Gaussians every 100 iterations, starting densification after 500 iterations. The total optimization process involves 10,000 steps. Following FSGS, pseudo views are enabled after 2,000 iterations. During training, we initialize the spherical harmonics (SH) degree at 0 and increment it by 1 every 1000 iterations, up to a maximum degree of 3, progressively refining the lighting representation. Simultaneously, opacity values are reset every 3000 iterations to eliminate low-opacity floaters by clamping all opacities to a maximum value of 0.05 using an inverse sigmoid function. The learning rates for position, SH coefficients, opacity, scaling, and rotation are set to 0.00016, 0.0025, 0.05, 0.005, and 0.001, respectively.


\end{document}